\documentclass[10pt,twocolumn,letterpaper]{article}
\usepackage[T1]{fontenc}
\usepackage[pagenumbers]{cvpr}

\usepackage{cuted} 

\usepackage{currfile} 

\usepackage{caption} 

\usepackage{microtype}

\usepackage{booktabs}
\usepackage{array}
\usepackage{multirow}
\usepackage{amsmath,amssymb,bm}
\usepackage{algorithm}
\usepackage{algpseudocode}
\usepackage{tikz}
\usetikzlibrary{positioning,arrows.meta,fit,calc,backgrounds}
\usepackage{xspace}
\usepackage{siunitx}
\usepackage[normalem]{ulem}
\usepackage{pifont}

\newcommand{\method}{\textsc{Clasp}\xspace}
\newcommand{\psd}{\textsc{Esp}\xspace}
\newcommand{\ccd}{\textsc{Ccd}\xspace}
\newcommand{\csp}{\textsc{Csp}\xspace}
\newcommand{\sast}{\textsc{Sast}\xspace}
\newcommand{\ske}{\textsc{Ske}\xspace}
\newcommand{\pcd}{\textsc{Pcd}\xspace}
\newcommand{\methodbase}{\textsc{Clasp}-Base\xspace}
\newcommand{\methodce}{\textsc{Clasp}\textsubscript{CE}\xspace}

\definecolor{qGood}{HTML}{0F7A3D}
\definecolor{qBad}{HTML}{C0392B}

\newcommand{\unk}{$\langle$unk$\rangle$}
\newcommand{\best}[1]{\textbf{#1}}
\newcommand{\second}[1]{\underline{#1}}

\newcommand{\pmstd}[1]{{\scriptsize$\pm$#1}}
\newcommand{\supp}[1]{Supp.~\ref{#1}}
\definecolor{cvprblue}{rgb}{0.21,0.49,0.74}
\usepackage[pagebackref,breaklinks,colorlinks,allcolors=cvprblue]{hyperref}

\def\paperID{*****}
\def\confName{CVPR}
\def\confYear{2027}

\title{\textsc{Clasp}: Deciding Structure and Claims Before Decoding\\for Whole-Slide Pathology Report Generation}
\author{
Ziaullah Khan$^{1}$\qquad Rashadul Islam Sumon$^{1}$\qquad Md Ariful Islam Mozumder$^{1}$\\
Duc-Son Pham$^{2}$\qquad Moon-Il Joo$^{1}$\qquad Hee-Cheol Kim$^{1}$\\[0.6em]
$^{1}$Digital Anti-aging Healthcare, Inje University, South Korea\\
$^{2}$Computing, Curtin University of Technology, Australia
}

\begin{document}
\maketitle
\begin{abstract}
Whole-slide pathology report generation has advanced through stronger slide encoders, retrieved knowledge and, recently,
slide-level multimodal LLMs. However, these generators leave every report-level choice to the decoder: how long the report
is, how it is organised, and which findings it states and where. On the standard PathText-BRCA benchmark, we find that
these choices dominate n-gram overlap, as reports of the query's hospital selected without learning approach the state of
the art, while decoders under-generate and fall back to the most frequent findings. We present \method, a
decide-then-decode generator that makes these choices explicitly, once per slide. Before decoding, an Exemplar Structure
Prior (ESP) takes the sentence plan and length from the reports of image-similar cases, a Calibrated Claim Decision (CCD)
selects the findings from calibrated slide posteriors, and a Claim-to-Sentence Placement (CSP) assigns each finding to a
planned sentence. A Plan-Conditioned Decoder (PCD), refined by structure-anchored sequence training, then writes the report
under these decisions. With 35M parameters, \method outperforms re-trained report generators and slide-level LLMs of up to
7B parameters, matches the strongest published results, and generalises to lung cancer from unseen hospitals. Its
decisions change what the reports state, not only their length, transfer to an LLM decoder and can withhold what H\&E
cannot show: withholding receptor status leaves 23\% fewer contradictions with clinical records than the same network
without decisions. Code and
models are available at
\mbox{\href{https://anonymous.4open.science/r/CVPR-2027-FFB8/}{\nolinkurl{anonymous.4open.science/r/CVPR-2027-FFB8}}}.
\end{abstract}

\section{Introduction}
\label{sec:intro}

\definecolor{tBad}{HTML}{C0392B}\definecolor{tGood}{HTML}{0F7A3D}\definecolor{tInk}{HTML}{2B2B2B}
\definecolor{tPanel}{HTML}{F2F2F0}\definecolor{tEsp}{HTML}{EB6834}\definecolor{tClm}{HTML}{1BAF7A}\definecolor{tDec}{HTML}{2A78D6}
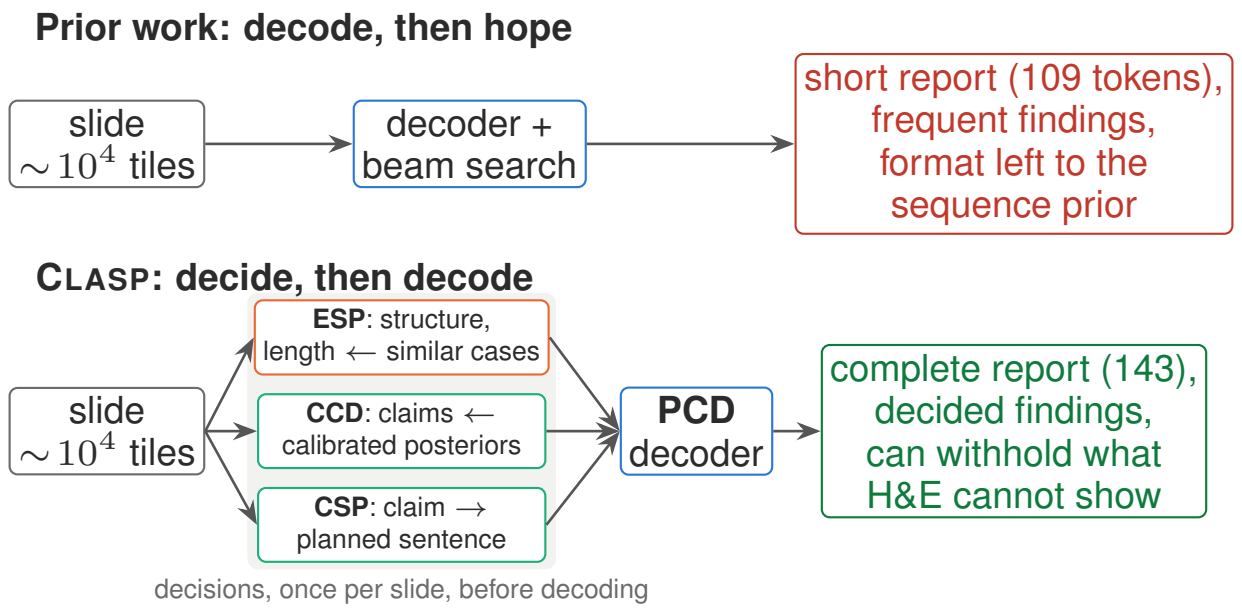
\begin{figure}[t]
\centering
\begin{tikzpicture}[x=1mm,y=1mm, font=\scriptsize\sffamily, text=tInk,
  box/.style={draw=tInk!70, rounded corners=1.5pt, fill=white, align=center, inner sep=1.6pt, line width=.4pt},
  arr/.style={-{Stealth[length=1.6mm]}, line width=.5pt, draw=tInk!80},
  ttl/.style={font=\scriptsize\sffamily\bfseries, anchor=west}]
\node[ttl] at (0,27.5) {Prior work: decode, then hope};
\node[box] (s1) at (6,20) {slide\\$\sim\!10^4$ tiles};
\node[box, draw=tDec] (d1) at (30,20) {decoder +\\beam search};
\node[box, draw=tBad, text=tBad] (r1) at (66,20) {short report (109 tokens),\\frequent findings,\\format left to the\\sequence prior};
\draw[arr] (s1) -- (d1); \draw[arr] (d1) -- (r1);
\node[ttl] at (0,11) {\textsc{Clasp}: decide, then decode};
\node[box] (s2) at (6,1) {slide\\$\sim\!10^4$ tiles};
\node[box, draw=tEsp, font=\tiny\sffamily, minimum width=19mm] (e) at (25.5,7.2) {\textbf{ESP}: structure,\\length $\leftarrow$ similar cases};
\node[box, draw=tClm, font=\tiny\sffamily, minimum width=19mm] (c) at (25.5,1) {\textbf{CCD}: claims $\leftarrow$\\calibrated posteriors};
\node[box, draw=tClm, font=\tiny\sffamily, minimum width=19mm] (p) at (25.5,-5.2) {\textbf{CSP}: claim $\rightarrow$\\planned sentence};
\node[box, draw=tDec] (d2) at (45,1) {\textbf{PCD}\\decoder};
\node[box, draw=tGood, text=tGood] (r2) at (66,1) {complete report (143),\\decided findings,\\can withhold what\\H\&E cannot show};
\draw[arr] (s2.east) -- (e.west); \draw[arr] (s2) -- (c); \draw[arr] (s2.east) -- (p.west);
\draw[arr] (e.east) -- (d2.west); \draw[arr] (c) -- (d2); \draw[arr] (p.east) -- (d2.west); \draw[arr] (d2) -- (r2);
\begin{scope}[on background layer]
\node[fill=tPanel, rounded corners=2pt, fit=(e)(c)(p), inner sep=1.2pt] {};
\end{scope}
\node[font=\tiny\sffamily, text=tInk!70] at (25.5,-9.6) {decisions, once per slide, before decoding};
\end{tikzpicture}
\caption{\textbf{Decide, then decode.} Report generators leave the report-level choices (length, organisation, which
findings and where) to beam search, which produces short reports of frequent findings (references: 149 tokens).
\method decides them once per slide (\psd, \ccd, \csp), and its \pcd realises them; a decision can also withhold what
the slide cannot show.}
\label{fig:teaser}
\end{figure}

A diagnostic pathology report condenses what a pathologist finds in the slides of a case (tumour type and grade,
in-situ components, margins, lymph nodes, receptor status) into a few paragraphs of structured prose. Generating such
reports from whole-slide images (WSIs) is a demanding vision-language problem: a slide is a bag of some $10^4$ tiles
with sparse diagnostic evidence, the report is long and follows the format of the reporting hospital, and paired data
are scarce. On the standard benchmark, PathText-BRCA~\cite{chen2024wsicaption}, progress has come from better evidence for
the decoder, through multiple-instance encoders~\cite{chen2024wsicaption}, hierarchical slide
representations~\cite{guo2024histgen}, retrieved report sentences~\cite{zhang2025bigen} and re-ranked
knowledge~\cite{chen2026ranger}, and from slide-level multimodal LLMs~\cite{tran2025histogpt,chen2025slidechat}.

As illustrated in Fig.~\ref{fig:teaser} (top), all of these generators leave the \emph{report-level} choices to
decoding. How long a report is and how it is organised, which findings it states and in which sentence, are settled
token by token by beam search over a maximum-likelihood decoder, which favours short outputs and frequent findings. We
find that these choices dominate what the benchmark measures (Sec.~\ref{sec:motivation}). Pathology foundation-model
embeddings reveal the reporting hospital~\cite{howard2021site}, and the most typical of 30 \emph{random} reports from the
query's hospital matches the published BLEU-4 of BiGen~\cite{zhang2025bigen}. Generators write about 110 tokens where
pathologists write 149, and the findings they state collapse to the most frequent ones, although their own slide
classifier recognises better ones.

This raises a simple question: rather than hoping that the decoder settles these choices, can we decide them? In this
paper, we answer it with \method (Claim- and Structure-Planned generation), a \emph{decide-then-decode} generator
(Fig.~\ref{fig:teaser}, bottom) that makes the report-level choices explicitly, once per slide, each from the source that
carries the information. The \emph{Exemplar Structure Prior} (\psd) takes the sentence plan and length budget from the
reports of image-similar training cases, typically of the same hospital. The \emph{Calibrated Claim Decision} (\ccd)
selects the findings from slide posteriors that are calibrated analytically. The \emph{Claim-to-Sentence Placement}
(\csp) assigns every finding to the planned sentence whose type states it most often. A \emph{Plan-Conditioned Decoder}
(\pcd) reads the three decisions at every step, together with the evidence of a \emph{Slide Knowledge Encoder} (\ske),
and is trained under them with cross-entropy and then with \emph{Structure-Anchored Sequence Training} (\sast). Deciding these
choices separates what the slide determines (findings) from what similar reports determine (format, length), so the
decoder only realises them.

This is a substantial departure from prior generators, from multiple-instance decoders to slide-level LLMs, which improve
what evidence reaches the decoder but not what the decoder decides. To the best of our knowledge, \method is the first
whole-slide report generator that decides the structure of a report, its findings and their placement before decoding.
\method is also controllable and portable. First, its decisions are explicit, so they can be inspected and overridden:
withholding a variable that H\&E cannot determine, such as receptor status, becomes a guarantee rather than a hope.
Second, they are discrete, so they transfer: given as text to a 7B LLM, they nearly double its agreement with the
patients' clinical records. Remarkably, with 35M parameters, \method outperforms re-trained slide-level LLMs of up to 7B parameters.

To summarize, we make the following contributions:
\begin{itemize}
\item We show that n-gram overlap on PathText-BRCA is dominated by report-level choices (the hospital's format, the
report length and the frequent findings) and that released slide-level LLMs have seen its test reports
(Secs.~\ref{sec:motivation} and~\ref{sec:setup}).
\item We introduce \method, a decide-then-decode generator whose report structure (\psd), findings (\ccd) and their
placement (\csp) are decided before decoding and realised by a plan-conditioned decoder trained with structure-anchored
sequence training (Sec.~\ref{sec:method}).
\item We demonstrate that \method outperforms re-trained report generators and slide-level LLMs, matches the strongest
published results and transfers to lung cancer from unseen hospitals, and that its explicit decisions transfer to an LLM
and, through explicit abstention, leave 23\% fewer contradictions with the clinical record than the same network without
decisions (Sec.~\ref{sec:exp}).
\end{itemize}
We release code and models, and we hope that explicit report-level decisions become a standard part of report
generation.

\section{Related Work}
\label{sec:related}

\noindent\textbf{Whole-slide report generation.} A whole-slide image is commonly treated as a bag of tile embeddings
from pathology foundation models~\cite{chen2024uni,uni2h2025,lu2024conch}, aggregated by multiple-instance
learning~\cite{lu2021clam}. WsiCaption~\cite{chen2024wsicaption} introduced the PathText benchmark together with MI-Gen, a
multiple-instance Transformer that writes the report from patch features, and HistGen~\cite{guo2024histgen} encodes the
slide hierarchically and aligns it with text through a cross-modal memory. Retrieval-based methods followed:
BiGen~\cite{zhang2025bigen} retrieves training-report sentences with PLIP~\cite{huang2023plip} embeddings,
RANGER~\cite{chen2026ranger} re-ranks retrieved knowledge and decodes with a sparse mixture of experts, and
SCOUT~\cite{singh2026scout} conditions the decoder on slide-level and concept streams. Despite these advances, all of
them improve \emph{what} evidence reaches the decoder and leave the report-level choices to decoding. In contrast,
\method decides the structure, the findings and their placement before the first word is written.

\noindent\textbf{Slide-level foundation models and multimodal LLMs.} Slide encoders pre-trained on large paired
corpora, such as PRISM~\cite{shaikovski2024prism} and TITAN~\cite{ding2025titan}, produce slide embeddings and generate
captions or reports, and multimodal LLMs such as HistoGPT~\cite{tran2025histogpt}, SlideChat~\cite{chen2025slidechat},
WSI-LLaVA~\cite{liang2025wsillava} and PathChat~\cite{lu2024pathchat} write reports or answer questions about slides
and regions. These models add capacity, but they still decode a report as one free-form sequence. In contrast,
\method's decisions are discrete and also improve such a decoder when given as text (Sec.~\ref{sec:ablation}).

\noindent\textbf{Planning, retrieval and length in text generation.} Radiology report generators draw on memories of
report patterns~\cite{chen2020r2gen,chen2021r2gencmn}, finding graphs~\cite{zhang2020kg} and retrieved
templates~\cite{li2018hrgr,li2019kerp}; PromptMRG~\cite{jin2024promptmrg} turns disease classifiers into prompts, and
ORGAN~\cite{hou2023organ} plans observations before writing. More broadly, content planning separates what to say from
how to say it~\cite{puduppully2019planning,krause2017paragraph}, and retrieved text usually enters a generator as
content~\cite{guu2018prototype,lewis2020rag}. Length is a known weakness of beam search under locally normalised models, which prefers
short outputs~\cite{murray2018length,stahlberg2019cat}; length normalisation~\cite{wu2016gnmt} re-scores finished
hypotheses, and length-controllable models~\cite{kikuchi2016length,takase2019lrpe,deng2020labert} condition on a given
length, which in report generation is unknown at test time. While similar in spirit to planned generation, \method
differs in two ways. First, it takes only the \emph{structure} and length of retrieved reports, whose content belongs to
another patient (Sec.~\ref{sec:analysis}), and decides every finding from the slide. Second, its structure signals
anchor self-critical training~\cite{rennie2017scst}, which otherwise drifts in length on long reports
(\supp{app:negative}).

\noindent\textbf{Evaluation and shortcuts.} Report generators are ranked by overlap metrics
(BLEU~\cite{papineni2002bleu}, METEOR~\cite{banerjee2005meteor}, ROUGE-L~\cite{lin2004rouge},
CIDEr~\cite{vedantam2015cider}); the REG~2025 challenge adds keyword and embedding similarity~\cite{lee2026reg}, and
PathReportEval scores clinical fields~\cite{singh2026pathreporteval}. Site signatures of TCGA slides inflate
classifiers~\cite{howard2021site}, a shortcut~\cite{geirhos2020shortcut}; we show that they also inflate report metrics,
and we evaluate against clinical records rather than with an extractor that also defines training labels.

\section{Method}
\label{sec:method}

We introduce \method, which first decides the report-level variables of a whole-slide image and then decodes the
report under them. Sec.~\ref{sec:motivation} defines the problem and motivates the design, Sec.~\ref{sec:overview} gives an
overview, and Secs.~\ref{sec:backbone} to~\ref{sec:abstain} describe the components.

\subsection{Problem Definition and Motivation}
\label{sec:motivation}
\label{sec:analysis}

\noindent\textbf{Problem definition.} A whole-slide image is represented by its tile embeddings
$\mathbf{X}=\{\mathbf{x}_i\}_{i=1}^{N}$, $\mathbf{x}_i\in\mathbb{R}^{d_v}$, and a training corpus
$\mathcal{D}=\{(\mathbf{X}^{(j)},y^{(j)})\}_{j=1}^{M}$ pairs slides with reports. We generate a report
$y=(y_1,\dots,y_T)$ that states the findings supported by $\mathbf{X}$ in the form a pathologist would write, without
access to the reference or to case metadata (hospital, specimen list). Everything derived from reports uses training
reports only, and retrieval never returns the query (notation: \supp{app:backbone}).

\begin{figure}[t]
\centering
\includegraphics[width=0.74\linewidth, trim=0 0 193 0, clip]{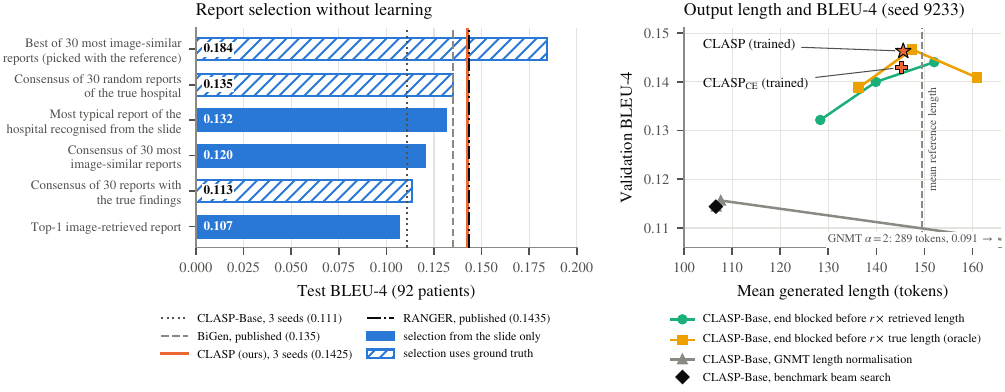}
\caption{\textbf{What BLEU-4 rewards on PathText-BRCA.} Test BLEU-4 of report \emph{selection} without learning
(hatched: uses ground truth). Reports of the query's hospital approach the state of the art; reports with its true
findings do not.}
\label{fig:benchmark}
\end{figure}

\noindent\textbf{Motivation.} Three properties of the benchmark, which hold on both splits, motivate \method (test values,
as in published scores; \supp{app:benchmark}). First, overlap is dominated by the hospital's reporting format, which the slide reveals: a linear
probe on the mean UNI2-h~\cite{uni2h2025} tile embedding identifies the site of 92.4\% of test slides (majority:
15.2\%), and the consensus~\cite{devlin2015nn} of 30 \emph{random} reports of the query's site scores 0.135 BLEU-4,
BiGen's published score~\cite{zhang2025bigen}, against 0.113 for the 30 reports that best match its \emph{true}
findings (Fig.~\ref{fig:benchmark}). Retrieved reports thus carry structure, not findings: their length is right
(144 vs.\ 149 tokens), but only 38.7\% of their findings hold for the query. Second, generators
under-generate: \methodbase, trained without decisions, writes 109 tokens against 149, and imposing the mean
length of the five most image-similar reports raises its validation BLEU-4 from 0.114 to 0.144. Third, its stated
findings (micro-F1 0.432) are about as accurate as writing the four most frequent findings in every report (0.403) and trail its
own slide classifier (0.499; Sec.~\ref{sec:claims}).

\subsection{Overview}
\label{sec:overview}

\begin{figure*}[t]
\centering
\includegraphics[width=\textwidth]{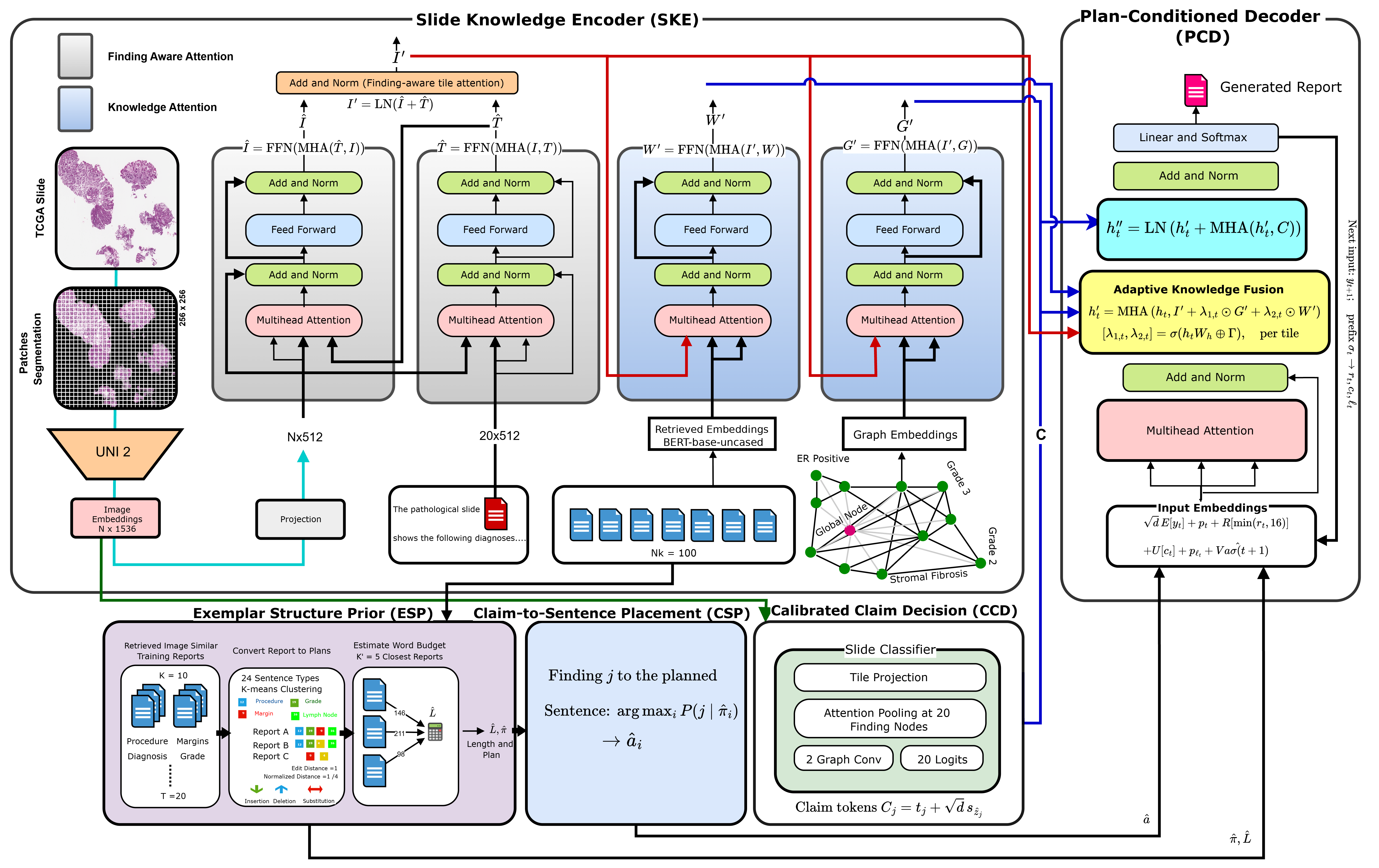}
\caption{\textbf{Overview of \method.} The Slide Knowledge Encoder (\ske) turns the UNI2-h tile embeddings into
finding-aware evidence $\mathbf{I}'$ and relates it to the reports of image-similar training cases ($\mathbf{W}'$) and to
a finding graph ($\mathbf{G}'$). Before decoding, the Exemplar Structure Prior (\psd) decides the sentence plan
$\hat{\bm\pi}$ and the length $\hat L$ from the most image-similar training reports, the Calibrated Claim Decision
(\ccd) decides the findings from calibrated slide posteriors and encodes them as claim tokens $\mathbf{C}$, and the
Claim-to-Sentence Placement (\csp) assigns each claim to a planned sentence ($\hat{\mathbf{a}}$). The Plan-Conditioned
Decoder (\pcd) reads the evidence by adaptive knowledge fusion, the claims by claim attention, and the remaining plan,
sentence type and length in its input embedding.}
\label{fig:arch}
\end{figure*}

Given the tile embeddings $\mathbf{X}$ of a slide, our goal is to generate its report $y$. As illustrated in
Fig.~\ref{fig:arch}, \method consists of a Slide Knowledge Encoder (\ske), three decision modules (\psd, \ccd and \csp)
and a Plan-Conditioned Decoder (\pcd). Our pipeline begins with the \ske, which turns the tiles into finding-aware
evidence (Sec.~\ref{sec:backbone}). Before decoding, the \psd decides the structure and length of the report, the \ccd
decides its findings and the \csp places each finding in a planned sentence (Secs.~\ref{sec:esp} to~\ref{sec:cplan}).
Given these decisions, the \pcd then writes the report. Formally, beam search approximates $\arg\max_y
p_\theta(y\mid\mathbf{X})$, in which the report-level variables $\omega=(\bm\pi,L,z,\mathbf{a})$ (structure, length,
findings and placement) stay implicit; \method replaces this single search by
\begin{equation}
\hat\omega=\big(\delta_{\mathrm{S}}(\mathbf{X}),\delta_{\mathrm{C}}(\mathbf{X}),\delta_{\mathrm{P}}\big),\;
\hat y=\arg\max_y p_\theta(y\mid\mathbf{X},\hat\omega),
\label{eq:principle}
\end{equation}
where each decision minimises its own risk with the evidence that determines it, and \sast fits the decoder to the
decisions (Sec.~\ref{sec:sast}).

\subsection{Slide Knowledge Encoder and Plan-Conditioned Decoder}
\label{sec:backbone}

We design the \ske to turn a slide into evidence that the report can describe. As a pathologist first looks for the
findings on a slide and then reads them in the light of earlier cases, the \ske has two steps, a finding-aware tile
attention and a knowledge attention. Both use multi-head attention $\mathrm{MHA}(\mathbf{A},\mathbf{B})$, in which the
rows of $\mathbf{A}$ query those of $\mathbf{B}$~\cite{vaswani2017attention}, and a two-layer perceptron $\mathrm{FFN}$,
each followed by dropout, a residual connection and layer normalisation ($\mathrm{LN}$).

\noindent\textbf{Inputs.} The \ske reads four sources, projected to $d{=}512$: the \emph{tiles}
$\mathbf{I}=\mathbf{X}\mathbf{W}_x\in\mathbb{R}^{N\times d}$; the \emph{topic bag} $\mathbf{T}\in\mathbb{R}^{N_T\times d}$,
word embeddings of the $N_T{=}20$ most frequent findings of the training reports; the \emph{retrieved reports}
$\mathbf{W}_{\mathrm{Pr}}\in\mathbb{R}^{N_K\times d}$, max-pooled BERT embeddings of the reports of the $N_K{=}100$ most
image-similar training cases; and the \emph{finding graph} $\mathbf{G}_{\mathrm{Pr}}$.

\noindent\textbf{Finding graph.} We design a finding graph so that related findings share evidence. Following
Zhang~\etal~\cite{zhang2020kg}, we build from the training corpus a universal graph whose nodes are the $N_T$ findings
and a global node. Where Zhang~\etal group chest findings by organ, we group breast findings by tissue compartment
(invasive tumour, ducts and lobules, benign parenchyma, axillary lymph nodes and surgical margins) and connect the
findings of a compartment by bidirectional edges; the global node is connected to every finding to add slide-level
context. Given a slide, each finding node is initialised from the tiles that resemble its finding and the global node
from all tiles, and two graph convolutions~\cite{kipf2017gcn} let each node gather its neighbours' evidence. The encoded
finding nodes, without the global node, form $\mathbf{G}_{\mathrm{Pr}}\in\mathbb{R}^{N_T\times d}$ (\supp{app:backbone}).

\noindent\textbf{Finding-aware tile attention.} A report names findings, so we first find which findings the slide shows
and where:\predisplaypenalty=0 
\begin{align}
\hat{\mathbf{T}}&=\mathrm{FFN}(\mathrm{MHA}(\mathbf{I},\mathbf{T})),\quad
\hat{\mathbf{I}}=\mathrm{FFN}(\mathrm{MHA}(\hat{\mathbf{T}},\mathbf{I})),\nonumber\\
\mathbf{I}'&=\mathrm{LN}(\hat{\mathbf{I}}+\hat{\mathbf{T}}).
\label{eq:tenc}
\end{align}\predisplaypenalty=10000
Each tile queries the topic bag, so row $i$ of $\hat{\mathbf{T}}\in\mathbb{R}^{N\times d}$ mixes the findings that tile
$i$ resembles; these finding mixtures then query the slide and gather the supporting tiles,
$\hat{\mathbf{I}}\in\mathbb{R}^{N\times d}$. We query with the tiles first because this filters out the findings that a
slide does not show~\cite{liu2021exploring}. As $\hat{\mathbf{I}}$ and $\hat{\mathbf{T}}$ are aligned tile by tile,
their sum $\mathbf{I}'$ describes every tile with its findings, a first reading of the slide.

\noindent\textbf{Knowledge attention.} The slide alone does not tell which findings occur together or how similar cases
were reported, so we let the finding-aware tiles query the graph and the retrieved reports,
\begin{align}
\mathbf{G}'&=\mathrm{FFN}(\mathrm{MHA}(\mathbf{I}',\mathbf{G}_{\mathrm{Pr}})),\nonumber\\
\mathbf{W}'&=\mathrm{FFN}(\mathrm{MHA}(\mathbf{I}',\mathbf{W}_{\mathrm{Pr}})),
\label{eq:kenc}
\end{align}
where $\mathbf{G}',\mathbf{W}'\in\mathbb{R}^{N\times d}$ hold the related findings and report content for every tile.

\noindent\textbf{Plan-Conditioned Decoder.} The \pcd is a three-layer Transformer decoder. At step $t$, masked
self-attention over the prefix gives $\mathbf{h}_t$, and \emph{adaptive knowledge fusion} reads the three sources,
\begin{equation}
\mathbf{h}'_t=\mathrm{MHA}\big(\mathbf{h}_t,\,\mathbf{I}'+\bm{\lambda}_{1,t}\odot\mathbf{G}'+\bm{\lambda}_{2,t}\odot\mathbf{W}'\big),
\label{eq:distil}
\end{equation}
where the sigmoid gates $\bm{\lambda}_{1,t},\bm{\lambda}_{2,t}\in(0,1)^N$, computed from $\mathbf{h}_t$ and the sources,
weigh the auxiliary knowledge per tile and per word, so that it is used only where it helps. In \methodbase both nearly
close (\supp{app:backbone}), and the retrieved reports, whose findings belong to other patients
(Sec.~\ref{sec:analysis}), enter \method through their structure, in the \psd. An FFN and a softmax layer
predict the next word, and the decisions add their inputs to the decoder (Eqs.~\eqref{eq:esp}, \eqref{eq:claimattn}
and~\eqref{eq:cplan}).

\subsection{Exemplar Structure Prior (ESP)}
\label{sec:esp}

The \psd decides the structure of the report, its sentence plan and length budget, from image-similar
training reports.

\noindent\textbf{Sentence types.} $k$-means on TF-IDF features clusters the training sentences into $C$ types
(\supp{app:config}), so a report $y$ has the plan $\bm{\pi}(y)=(\pi_1,\dots,\pi_S)$.

\noindent\textbf{Structure-conditioned decoding.} Let $\sigma_t$ be the number of sentences closed in the prefix
$y_{\le t}$. Given a structure $(\bm{\pi},L)$, the input of the \pcd at step $t$ is
\begin{equation}
\mathbf{e}_t=\sqrt{d}\,\mathbf{E}[y_t]+\mathbf{p}_t+\mathbf{R}[\min(r_t,R)]+\mathbf{U}[c_t]+\mathbf{p}_{\ell_t},
\label{eq:esp}
\end{equation}
where $r_t=\max(S-\sigma_t,0)$ are the remaining sentences, $c_t=\pi_{\sigma_t+1}$ the type of the sentence being written
(an \textsc{end} type after the plan), $\ell_t=\max(L-t,0)$ the remaining tokens, $\mathbf{E}$ the word embedding,
$\mathbf{p}$ the sinusoidal position code (re-used for the remaining length~\cite{takase2019lrpe}) and
$\mathbf{R},\mathbf{U}$ learned tables ($R{=}16$). Since $\sigma_t$ is a function of the prefix, Eq.~\eqref{eq:esp} is
exact under incremental beam search. In training the structure is the target's own, $(\bm{\pi}(y),|y|)$.

\noindent\textbf{Structure at test time.} The structure is taken from the $K$ training cases $\mathcal{N}_K(\mathbf{X})$
with the most similar mean tile embedding:
\begin{align}
\hat{\bm{\pi}}&=\arg\min_{\bm{\pi}_a\in\Pi_K}\sum_{\bm{\pi}_b\in\Pi_K\setminus\{\bm{\pi}_a\}}d_{\mathrm{ed}}(\bm{\pi}_a,\bm{\pi}_b),\nonumber\\
\hat L&=\frac{1}{K'}\sum_{j\in\mathcal{N}_{K'}(\mathbf{X})}|y^{(j)}|,
\label{eq:prior}
\end{align}
with $\Pi_K$ the plans of $\mathcal{N}_K(\mathbf{X})$, $d_{\mathrm{ed}}$ the length-normalised edit distance, $K{=}10$ and
$K'{=}5$: a structure written for a similar case, typically at the same hospital. A report that stops early
is inconsistent with the remaining-sentence and remaining-token signals, so the mode is a complete report.

\subsection{Calibrated Claim Decision (CCD)}
\label{sec:claims}

The \ccd decides which findings the report states, from the slide alone. The slide-level classifier $f_\phi$ (tile projection, graph and multi-label head) is trained with a binary
cross-entropy that weights the positives of finding $j$ by $w_j=n^-_j/n^+_j$. The minimiser of this loss has logit
$s_j=\operatorname{logit}p_j+\log w_j$ with $p_j=P(z_j{=}1\mid\mathbf{X})$, so the class weighting is undone
analytically, without fitting a calibrator, and the posterior is thresholded:
\begin{equation}
\tilde p_j(\mathbf{X})=\sigma\big(s_j(\mathbf{X})-\log w_j\big),\qquad \hat z_j=\mathbb{1}\big[\tilde p_j(\mathbf{X})\ge\tau\big].
\label{eq:claims}
\end{equation}
For calibrated posteriors the micro-F1-optimal decision is a single threshold~\cite{lipton2014thresholding}; one global $\tau$
is chosen on validation micro-F1 before the generator is trained. Each finding becomes a claim token
$\mathbf{c}_j=\mathbf{t}_j+\sqrt{d}\,\mathbf{s}_{\hat z_j}$ (topic embedding plus a learned present/absent state), which
every \pcd layer reads after the knowledge sources,
\begin{equation}
\mathbf{h}''_t=\mathrm{LN}\big(\mathbf{h}'_t+\mathrm{Drop}(\mathrm{MHA}(\mathbf{h}'_t,\mathbf{C}))\big),\qquad
\mathbf{C}=[\mathbf{c}_1;\dots;\mathbf{c}_{N_T}].
\label{eq:claimattn}
\end{equation}
During cross-entropy training the claims are the target's own findings; at test time they are $\hat z(\mathbf{X})$.

\subsection{Claim-to-Sentence Placement (CSP)}
\label{sec:cplan}

The \csp decides in which sentence each finding is stated. A finding is usually stated in one kind of sentence. Let $P(j\mid k)$ be the fraction of
training sentences of type $k$ that state finding $j$. Every claim is placed in the planned sentence whose type states
it most often (the earliest on ties),
\begin{equation}
\hat{\mathbf{a}}_i=\sum_{j:\,\hat z_j=1}\mathbb{1}\Big[i=\arg\max_{i'\le S}P\big(j\mid\hat\pi_{i'}\big)\Big]\,\mathbf{u}_j,
\qquad i=1,\dots,S,
\label{eq:cplan}
\end{equation}
($\mathbf{u}_j$ the $j$-th unit vector; during training, the findings each target sentence states), and the decoder input
of Eq.~\eqref{eq:esp} receives the findings of the sentence being written, $\mathbf{V}\hat{\mathbf{a}}_{\sigma_t+1}$ (\supp{app:backbone}).

\subsection{Structure-Anchored Sequence Training (SAST)}
\label{sec:sast}

\noindent\textbf{Training stages.} \method is trained in three stages. The tile projection and graph of the \ske are
first trained with the multi-label head of the \ccd (Sec.~\ref{sec:claims}), which keeps this model frozen; \ske and \pcd are then trained with
cross-entropy under the target's own structure and findings; finally, \sast fits the decoder to the test-time decisions.

\noindent\textbf{Sequence-level objective.} Cross-entropy training conditions the decoder on the target's \emph{own} structure and findings, whereas at test time it
receives the retrieved structure and the decided claims, and it never sees its own samples. Optimising a report-level
reward on samples closes both gaps, but for long reports such training is fragile (\supp{app:negative}); the structure
signals of Eq.~\eqref{eq:esp} anchor it. Starting from the cross-entropy model, we minimise the negative expected reward of reports sampled under the
test-time decisions,
\begin{equation}
\mathcal{L}_{\mathrm{seq}}(\theta)=-\,\mathbb{E}_{y\sim p_\theta(\cdot\mid\mathbf{X},\hat{\bm\pi},\hat L,\hat z,\hat{\mathbf{a}})}\big[r(y,y^\star)\big],
\label{eq:sast}
\end{equation}
where $y^\star$ is the reference and $r=\mathrm{BLEU\text{-}4}+\mathrm{METEOR}+\mathrm{ROUGE\text{-}L}$ sums the
report-level versions of the three metric families, so that the decoder cannot trade one for another. The gradient uses
$K_s{=}5$ samples per slide and the leave-one-out baseline~\cite{kool2019buy}. With the anchor, sampled reports keep
the planned length (139 to 141 tokens, against 181 to 187 without it). \sast adds no parameters.

\subsection{Explicit Abstention}
\label{sec:abstain}

A decision made before decoding can also withhold content. For a set of variables that the slide cannot
determine (here receptor status, which is measured by immunohistochemistry rather than on H\&E), \method (i) removes
from the \psd plan the sentence types whose training sentences mention the variables in at least 90\% of cases, shrinking the
budget by their share, (ii) withdraws the \ccd claims of the corresponding findings, and (iii) deletes after decoding every generated
sentence that the type model assigns to a withheld type. Step (iii) guarantees that no withheld type is emitted; steps (i) and (ii)
keep the rest of the report coherent. Nothing is trained or tuned, and the variables are fixed by clinical definition
(\supp{app:abstain}).

\section{Experiments}
\label{sec:exp}
\label{sec:setup}

This section compares \method with published and re-trained methods (Sec.~\ref{sec:sota}), measures its clinical
fidelity (Sec.~\ref{sec:clinical}) and ablates its components (Sec.~\ref{sec:ablation}).

\subsection{Datasets and Metrics}

\noindent\textbf{Datasets.} PathText-BRCA~\cite{chen2024wsicaption} pairs TCGA-BRCA diagnostic
slides~\cite{tcga2012brca} with their cleaned pathology reports~\cite{kefeli2024tcgareports}. We use the patient-level split of BiGen~\cite{zhang2025bigen} (also
used by RANGER~\cite{chen2026ranger}) with one slide per patient. UNI2-h~\cite{uni2h2025} embeddings of all tissue
tiles~\cite{w8yi2026features} exist for 969 of the 977 patients, which leaves 792/85/92 patients (\supp{app:config}).

\noindent\textbf{Metrics.} Corpus BLEU-1 to BLEU-4, METEOR, ROUGE-L and CIDEr with the benchmark's scorers. An
\emph{independent clinical evaluation} scores reports against the TCGA-BRCA clinical records~\cite{liu2018tcgacdr}, which play no role in training or selection:
histological type, estrogen, progesterone and HER2 receptor status, lymph-node and margin status, and tumour size. A
reader, Llama-3.1-8B-Instruct~\cite{grattafiori2024llama3}, answers one multiple-choice question per report and variable
(``not stated'' is an option) without seeing the truth. We report the macro-F1 over the truth classes and the
\emph{contradiction rate}, the share of patients whose report contradicts the record; on reference reports that state a
variable, the reader agrees with the record in 76 to 95\% of cases (\supp{app:clinical}). We also report
CRQS~\cite{singh2026pathreporteval}; the REG~2025 score~\cite{lee2026reg} and BERTScore~\cite{zhang2020bertscore} are in
\supp{app:modern}.

\subsection{Baselines and Implementation Details}

\noindent\textbf{Protocol.} Every model is selected on validation BLEU-4 (the benchmark's monitor) and decodes the test
split once; every design choice was made on validation by a rule fixed before the result was known. We report the mean
and standard deviation over three seeds and test differences with a paired bootstrap over test patients (10{,}000
resamples, seed-averaged systems).

\noindent\textbf{Baselines.} \emph{Published}: quoted from RANGER~\cite{chen2026ranger}; RANGER's code is not public, and BiGen needs PLIP features of the same patches; the public split of SCOUT's code~\cite{singh2026scout,singh2026pathreporteval} trains on 75 of our 92 test patients (\supp{app:baselines}).
\emph{Re-trained report generators}: MI-Gen~\cite{chen2024wsicaption} and HistGen~\cite{guo2024histgen} with their public
code (\supp{app:baselines}). \emph{Re-trained slide-level LLMs}, since the released SlideChat and WSI-LLaVA contain 75 and all 92 test
reports in their training data (\supp{app:llm}): HistoGPT-1B~\cite{tran2025histogpt}, fully fine-tuned; the SlideChat architecture~\cite{chen2025slidechat} (CONCH~\cite{lu2024conch} patch features, LongNet slide encoder,
Qwen2.5-7B-Instruct~\cite{yang2024qwen25} with LoRA~\cite{hu2022lora}); TITAN + LLM, the TITAN slide
embedding~\cite{ding2025titan} as prefix tokens of the same LLM; and the LLM without visual input. \emph{Retrieval
controls without learning}: the most image-similar training report, the $k$-NN consensus of the 30 most
similar~\cite{devlin2015nn}, and a \emph{hospital consensus}, the most typical report of the site of the most
image-similar slide. \emph{Our variants}: \methodbase, \ske and \pcd trained without decisions and without \sast; a
\emph{length control}, \methodbase decoded with the end token blocked before the \psd budget; \methodce, \psd, \ccd and
\csp with cross-entropy training only; and \method.

\noindent\textbf{Implementation details.} 8 heads, dropout 0.1; at most $10^4$ tiles per slide; beam 3 with the benchmark's trigram penalty; $C{=}24$ sentence types; claim
thresholds $\tau=0.17, 0.19, 0.27$ (one slide-level classifier per seed, chosen on validation). Training takes 1 to 2\,h plus 66 to
95\,min of \sast per seed on one RTX A5000 (\supp{app:config}); the decisions add 5.6M parameters and 3\% compute
(\supp{app:backbone}).

\begin{table*}[t]
\caption{\textbf{Comparison on PathText-BRCA (test).} Top: published single runs quoted from~\cite{chen2026ranger} (UNI at
10$\times$, 93 test patients). Bottom: run under our protocol (UNI2-h, 92 test patients, validation-selected checkpoints,
test decoded once); mean \pmstd{std} over three seeds (MI-Gen, HistGen: one seed; retrieval: deterministic). EMA:
weight averaging, selected on validation (\supp{app:ema}); receptors withheld: Sec.~\ref{sec:abstain}. F1 / Contr.:
macro-F1 and contradiction rate against the TCGA clinical records. CRQS: PathReportEval~\cite{singh2026pathreporteval}.
\best{Bold}/\second{underline}: best/second among the methods run here; $^{*}$/$^{\dagger}$: \method better/worse,
$p<0.05$ (one-sided paired bootstrap over test patients).}
\label{tab:main}
\centering\footnotesize
\setlength{\tabcolsep}{3pt}
\resizebox{\textwidth}{!}{%
\begin{tabular}{@{}lcccccccccc@{}}
\toprule
 & \multicolumn{7}{c}{Benchmark metrics} & \multicolumn{2}{c}{TCGA records} & \\
\cmidrule(lr){2-8}\cmidrule(lr){9-10}
Method & B-1 & B-2 & B-3 & B-4 & MTR & R-L & CIDEr & F1 & Contr.$\downarrow$ & CRQS\\
\midrule
\multicolumn{11}{@{}l}{\emph{Published (quoted)}}\\
CNN-RNN & 0.371 & 0.185 & 0.089 & 0.043 & 0.143 & 0.239 & n/a & n/a & n/a & n/a\\
att-LSTM & 0.371 & 0.191 & 0.094 & 0.048 & 0.142 & 0.238 & n/a & n/a & n/a & n/a\\
Transformer & 0.389 & 0.246 & 0.157 & 0.103 & 0.158 & 0.257 & n/a & n/a & n/a & n/a\\
R2Gen & 0.378 & 0.243 & 0.160 & 0.107 & 0.179 & 0.279 & n/a & n/a & n/a & n/a\\
R2GenCMN & 0.396 & 0.254 & 0.164 & 0.110 & 0.163 & 0.279 & n/a & n/a & n/a & n/a\\
MI-Gen & 0.416 & 0.267 & 0.174 & 0.115 & 0.165 & 0.270 & n/a & n/a & n/a & n/a\\
HistGen & 0.422 & 0.272 & 0.177 & 0.118 & 0.169 & 0.277 & n/a & n/a & n/a & n/a\\
BiGen & 0.450 & 0.296 & 0.196 & 0.135 & 0.180 & 0.293 & n/a & n/a & n/a & n/a\\
RANGER & 0.4598 & 0.3044 & 0.2036 & 0.1435 & 0.1883 & 0.3038 & n/a & n/a & n/a & n/a\\
\midrule
\multicolumn{11}{@{}l}{\emph{Run under our protocol}}\\
MI-Gen~\cite{chen2024wsicaption} & 0.406$^{*}$ & 0.271$^{*}$ & 0.184$^{*}$ & 0.1265$^{*}$ & 0.157$^{*}$ & 0.295 & 0.076 & 0.259$^{*}$ & \best{0.108}$^{\dagger}$ & 0.259\\
HistGen~\cite{guo2024histgen} & 0.293$^{*}$ & 0.194$^{*}$ & 0.129$^{*}$ & 0.0866$^{*}$ & 0.130$^{*}$ & 0.273$^{*}$ & 0.049$^{*}$ & 0.201$^{*}$ & 0.161$^{\dagger}$ & 0.236\\
HistoGPT-1B~\cite{tran2025histogpt} & 0.312\pmstd{0.021}$^{*}$ & 0.198\pmstd{0.010}$^{*}$ & 0.127\pmstd{0.004}$^{*}$ & 0.0823\pmstd{0.0010}$^{*}$ & 0.136\pmstd{0.003}$^{*}$ & 0.267\pmstd{0.006}$^{*}$ & 0.052\pmstd{0.011}$^{*}$ & 0.265\pmstd{0.036}$^{*}$ & 0.191\pmstd{0.038}$^{\dagger}$ & 0.157\pmstd{0.064}$^{*}$\\
SlideChat arch.~\cite{chen2025slidechat} (7B) & 0.327\pmstd{0.077}$^{*}$ & 0.203\pmstd{0.048}$^{*}$ & 0.127\pmstd{0.028}$^{*}$ & 0.0791\pmstd{0.0162}$^{*}$ & 0.137\pmstd{0.019}$^{*}$ & 0.248\pmstd{0.007}$^{*}$ & 0.029\pmstd{0.004}$^{*}$ & 0.209\pmstd{0.071}$^{*}$ & 0.197\pmstd{0.081}$^{\dagger}$ & 0.185\pmstd{0.073}$^{*}$\\
TITAN~\cite{ding2025titan} + LLM (7B) & 0.340\pmstd{0.012}$^{*}$ & 0.208\pmstd{0.012}$^{*}$ & 0.127\pmstd{0.011}$^{*}$ & 0.0780\pmstd{0.0086}$^{*}$ & 0.140\pmstd{0.002}$^{*}$ & 0.251\pmstd{0.003}$^{*}$ & 0.039\pmstd{0.003}$^{*}$ & 0.175\pmstd{0.034}$^{*}$ & 0.152\pmstd{0.041}$^{\dagger}$ & 0.207\pmstd{0.030}$^{*}$\\
\quad + \method decisions as text & 0.405\pmstd{0.002}$^{*}$ & 0.244\pmstd{0.004}$^{*}$ & 0.146\pmstd{0.004}$^{*}$ & 0.0878\pmstd{0.0025}$^{*}$ & 0.161\pmstd{0.001}$^{*}$ & 0.260\pmstd{0.002}$^{*}$ & 0.076\pmstd{0.013} & 0.324\pmstd{0.010} & 0.210\pmstd{0.019} & 0.248\pmstd{0.034}\\
LLM without visual input (7B) & 0.327\pmstd{0.066}$^{*}$ & 0.201\pmstd{0.043}$^{*}$ & 0.126\pmstd{0.028}$^{*}$ & 0.0790\pmstd{0.0177}$^{*}$ & 0.136\pmstd{0.019}$^{*}$ & 0.246\pmstd{0.006}$^{*}$ & 0.054\pmstd{0.009}$^{*}$ & 0.233\pmstd{0.094}$^{*}$ & 0.291\pmstd{0.171}$^{*}$ & 0.188\pmstd{0.046}$^{*}$\\
Most similar training report & 0.439 & 0.267$^{*}$ & 0.167$^{*}$ & 0.1068$^{*}$ & 0.182$^{\dagger}$ & 0.281$^{*}$ & 0.062 & 0.305 & 0.163$^{\dagger}$ & 0.232\\
Consensus of 30 similar~\cite{devlin2015nn} & 0.445 & 0.280$^{*}$ & 0.179$^{*}$ & 0.1201$^{*}$ & \second{0.185}$^{\dagger}$ & 0.286$^{*}$ & 0.088 & 0.327 & 0.227 & 0.245\\
Hospital consensus & \best{0.468} & 0.301 & 0.195$^{*}$ & 0.1315$^{*}$ & \best{0.199}$^{\dagger}$ & \second{0.302} & 0.084 & \best{0.337} & 0.267 & 0.229\\
\midrule
\methodbase & 0.363\pmstd{0.020}$^{*}$ & 0.243\pmstd{0.013}$^{*}$ & 0.164\pmstd{0.008}$^{*}$ & 0.1106\pmstd{0.0058}$^{*}$ & 0.150\pmstd{0.009}$^{*}$ & 0.287\pmstd{0.013}$^{*}$ & 0.112\pmstd{0.024} & 0.293\pmstd{0.057}$^{*}$ & 0.169\pmstd{0.058}$^{\dagger}$ & 0.266\pmstd{0.071}\\
\methodbase + length control & 0.453\pmstd{0.019} & 0.304\pmstd{0.011} & 0.205\pmstd{0.007}$^{*}$ & 0.1378\pmstd{0.0055} & 0.176\pmstd{0.012} & 0.288\pmstd{0.013}$^{*}$ & 0.104\pmstd{0.017} & \second{0.330}\pmstd{0.060} & 0.189\pmstd{0.054}$^{\dagger}$ & 0.260\pmstd{0.073}\\
\midrule
\methodbase + \psd & 0.452\pmstd{0.008} & 0.308\pmstd{0.008} & 0.209\pmstd{0.008} & 0.1416\pmstd{0.0065} & 0.174\pmstd{0.003} & 0.293\pmstd{0.007}$^{*}$ & \best{0.126}\pmstd{0.010}$^{\dagger}$ & 0.314\pmstd{0.007} & 0.170\pmstd{0.021}$^{\dagger}$ & 0.244\pmstd{0.037}\\
\quad + \ccd & 0.448\pmstd{0.004} & 0.304\pmstd{0.005}$^{*}$ & 0.205\pmstd{0.004}$^{*}$ & 0.1350\pmstd{0.0044}$^{*}$ & 0.173\pmstd{0.002} & 0.287\pmstd{0.003}$^{*}$ & \second{0.121}\pmstd{0.006} & 0.289\pmstd{0.044}$^{*}$ & 0.180\pmstd{0.022}$^{\dagger}$ & 0.245\pmstd{0.023}\\
\quad + \csp = \methodce & 0.447\pmstd{0.003} & 0.306\pmstd{0.001} & 0.208\pmstd{0.003}$^{*}$ & 0.1383\pmstd{0.0042} & 0.172\pmstd{0.003} & 0.289\pmstd{0.006}$^{*}$ & \best{0.126}\pmstd{0.004}$^{\dagger}$ & 0.313\pmstd{0.048} & 0.185\pmstd{0.010}$^{\dagger}$ & \best{0.280}\pmstd{0.009}\\
\quad + \sast = \textbf{\method (ours)} & 0.450\pmstd{0.007} & \second{0.310}\pmstd{0.003} & \second{0.214}\pmstd{0.003} & \second{0.1425}\pmstd{0.0026} & 0.174\pmstd{0.002} & 0.301\pmstd{0.004} & 0.113\pmstd{0.014} & 0.328\pmstd{0.039} & 0.232\pmstd{0.059} & \second{0.269}\pmstd{0.030}\\
\quad \method-EMA (validation-selected) & \second{0.454}\pmstd{0.006}$^{\dagger}$ & \best{0.315}\pmstd{0.003}$^{\dagger}$ & \best{0.217}\pmstd{0.003} & \best{0.1445}\pmstd{0.0033} & 0.175\pmstd{0.002} & \best{0.303}\pmstd{0.002} & \second{0.121}\pmstd{0.024} & 0.279\pmstd{0.020}$^{*}$ & 0.209\pmstd{0.056}$^{\dagger}$ & 0.267\pmstd{0.029}\\
\quad \method, receptor status withheld & 0.405\pmstd{0.011}$^{*}$ & 0.279\pmstd{0.007}$^{*}$ & 0.191\pmstd{0.007}$^{*}$ & 0.1265\pmstd{0.0055}$^{*}$ & 0.162\pmstd{0.002}$^{*}$ & 0.298\pmstd{0.003}$^{*}$ & 0.073\pmstd{0.011}$^{*}$ & 0.222\pmstd{0.024}$^{*}$ & \second{0.130}\pmstd{0.026}$^{\dagger}$ & n/a\\
\bottomrule
\end{tabular}}
\end{table*}

\definecolor{wSa}{HTML}{2a78d6}\definecolor{wSb}{HTML}{eb6834}\definecolor{wSc}{HTML}{1baf7a}
\definecolor{wKn}{HTML}{D62828}\definecolor{wMuted}{HTML}{8A8984}
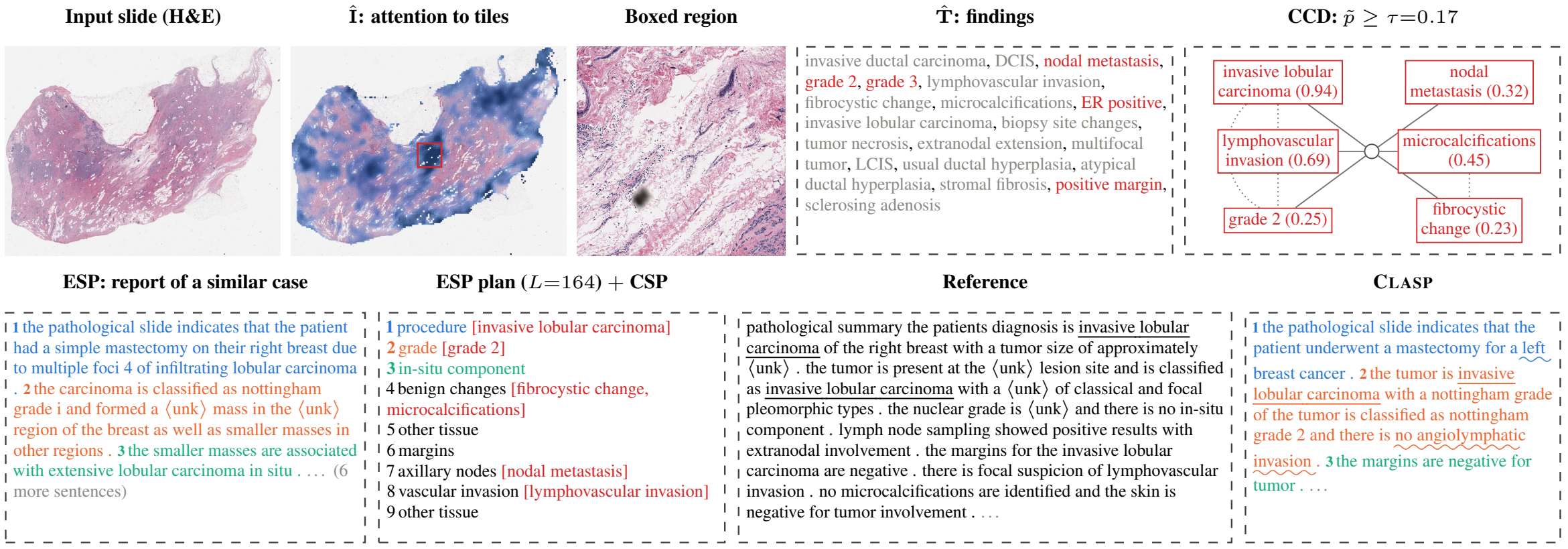
\begin{figure*}[t]
\newcommand{\wtitle}[1]{{\fontsize{6.4}{7.4}\selectfont\textbf{#1}}\\[1pt]}
\newcommand{\wframe}[3]{\tikz[baseline={([yshift=-5.5pt]b.north)}]\node[draw=black!75,dashed,line width=.5pt,inner sep=2.5pt](b)
  {\parbox[t][#2cm][t]{\dimexpr#1cm-5.5pt}{\fontsize{5.6}{6.55}\selectfont\raggedright#3}};}
\newcommand{\wpanel}[4]{\begin{minipage}[t]{#1cm}\centering\wtitle{#3}\wframe{#1}{#2}{#4}\end{minipage}}
\newcommand{\wimage}[3]{\begin{minipage}[t]{#1cm}\centering\wtitle{#2}\tikz[baseline={([yshift=-5.5pt]b.north)}]\node[inner sep=0pt](b){\includegraphics[width=#1cm,height=2.344cm]{#3}};\end{minipage}}
\centering
\resizebox{0.88\textwidth}{!}{\begin{minipage}{\textwidth}
\wimage{3.09}{Input slide (H\&E)}{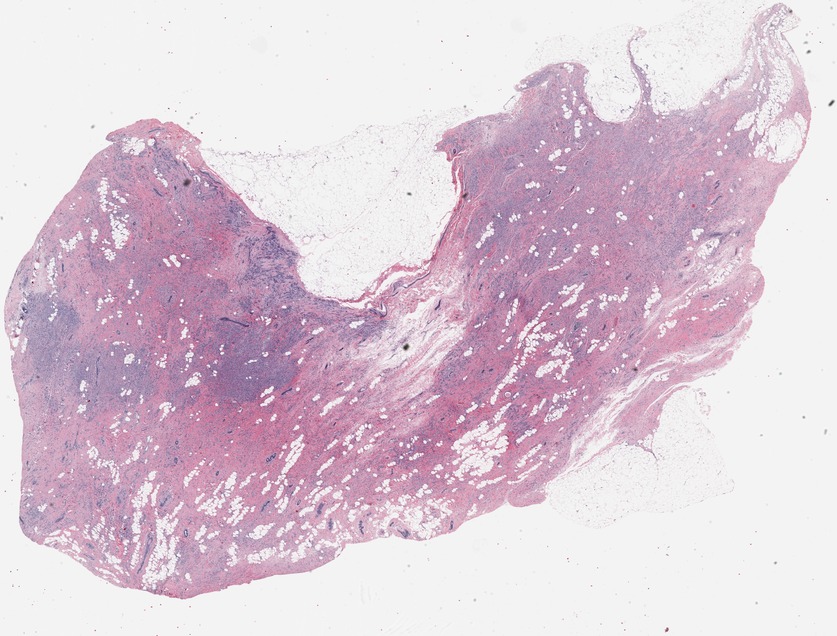}\hfill
\wimage{3.09}{$\hat{\mathbf{I}}$: attention to tiles}{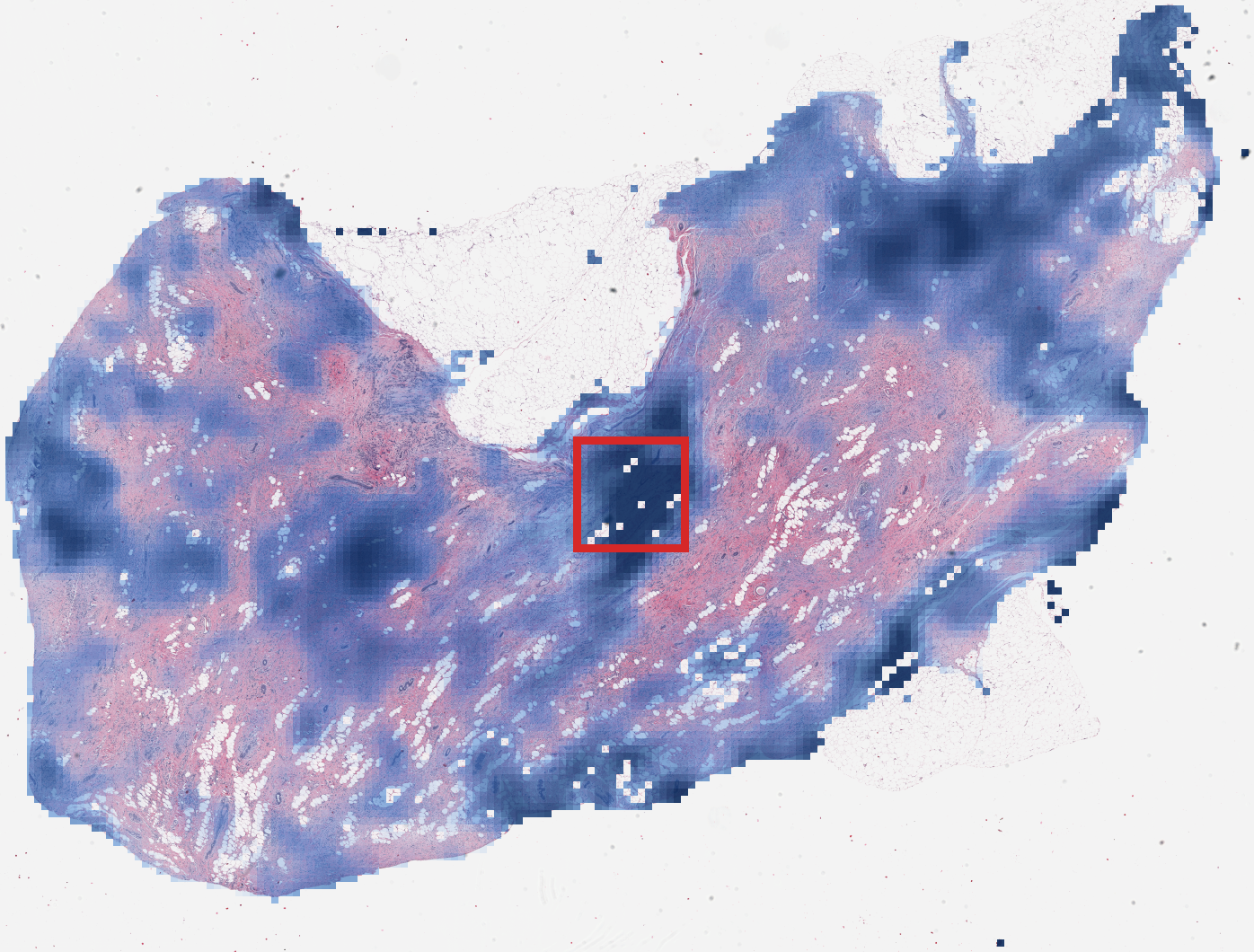}\hfill
\wimage{2.34}{Boxed region}{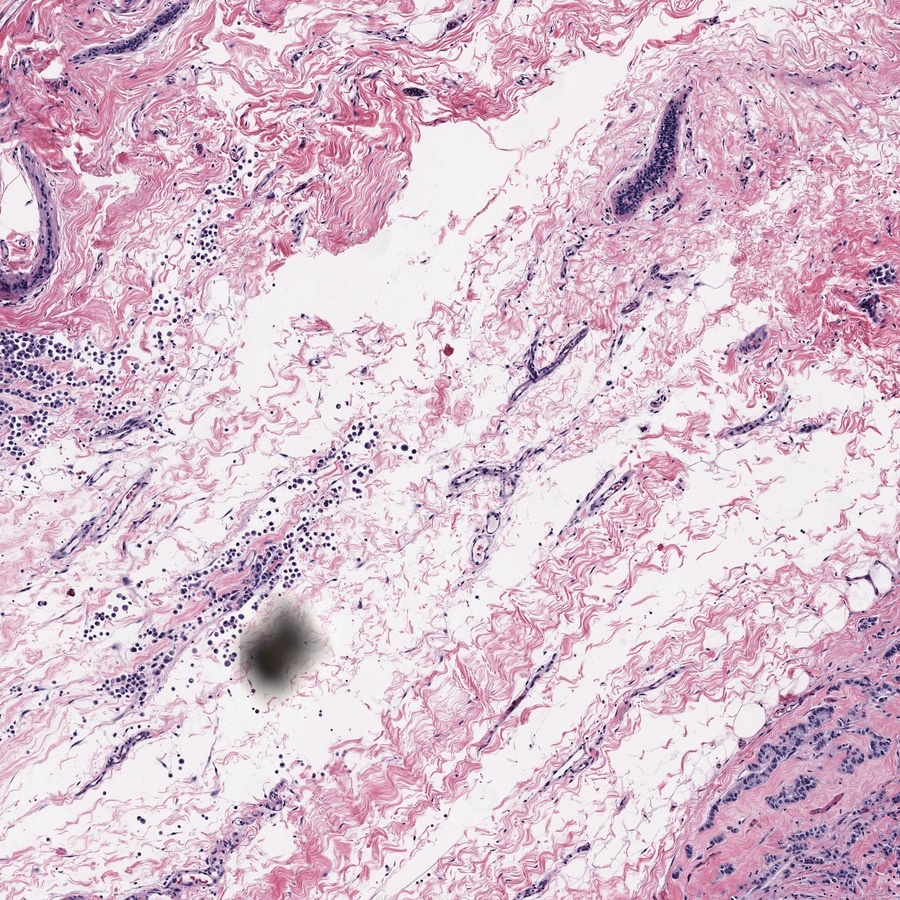}\hfill
\wpanel{4.23}{2.15}{$\hat{\mathbf{T}}$: findings}{\textcolor{wMuted}{invasive ductal carcinoma}, \textcolor{wMuted}{DCIS}, \textcolor{wKn}{nodal metastasis}, \textcolor{wKn}{grade 2}, \textcolor{wKn}{grade 3}, \textcolor{wMuted}{lymphovascular invasion}, \textcolor{wMuted}{fibrocystic change}, \textcolor{wMuted}{microcalcifications}, \textcolor{wKn}{ER positive}, \textcolor{wMuted}{invasive lobular carcinoma}, \textcolor{wMuted}{biopsy site changes}, \textcolor{wMuted}{tumor necrosis}, \textcolor{wMuted}{extranodal extension}, \textcolor{wMuted}{multifocal tumor}, \textcolor{wMuted}{LCIS}, \textcolor{wMuted}{usual ductal hyperplasia}, \textcolor{wMuted}{atypical ductal hyperplasia}, \textcolor{wMuted}{stromal fibrosis}, \textcolor{wKn}{positive margin}, \textcolor{wMuted}{sclerosing adenosis}}\hfill
\begin{minipage}[t]{4.23cm}\centering\wtitle{CCD: $\tilde p\ge\tau{=}0.17$}
\begin{tikzpicture}[x=1cm,y=1cm,baseline={([yshift=-5.5pt]frame.north)},cl/.style={draw=wKn,text=wKn,font=\fontsize{5.6}{6.2}\selectfont,align=center,inner sep=1.4pt,fill=white},gr/.style={draw=black!55,line width=.4pt}]
\node[draw=black!75,dashed,line width=.5pt,minimum width=4.210cm,minimum height=2.326cm,anchor=south west,inner sep=0] (frame) at (0,0) {};
\node[circle,draw=black!70,fill=white,inner sep=1.6pt] (g) at (2.105,1.163) {};
\node[cl] (n9) at (1.053,1.938) {invasive lobular\\carcinoma (0.94)};
\node[cl] (n5) at (1.053,1.163) {lymphovascular\\invasion (0.69)};
\node[cl] (n3) at (1.053,0.388) {grade 2 (0.25)};
\node[cl] (n2) at (3.200,1.938) {nodal\\metastasis (0.32)};
\node[cl] (n7) at (3.200,1.163) {microcalcifications\\(0.45)};
\node[cl] (n6) at (3.200,0.388) {fibrocystic\\change (0.23)};
\begin{scope}[on background layer]
\draw[gr] (g) -- (n9);
\draw[gr] (g) -- (n5);
\draw[gr] (g) -- (n3);
\draw[gr] (g) -- (n2);
\draw[gr] (g) -- (n7);
\draw[gr] (g) -- (n6);
\draw[gr, densely dotted] (n9) -- (n5);
\draw[gr, densely dotted] (n9) to[bend right=55] (n3);
\draw[gr, densely dotted] (n5) -- (n3);
\draw[gr, densely dotted] (n7) -- (n6);
\end{scope}
\end{tikzpicture}
\end{minipage}\\[5pt]
\wpanel{4.05}{2.4}{ESP: report of a similar case}{{\color{wSa}\textbf{\scalebox{.85}{1}}\,the pathological slide indicates that the patient had a simple mastectomy on their right breast due to multiple foci 4 of infiltrating lobular carcinoma .} {\color{wSb}\textbf{\scalebox{.85}{2}}\,the carcinoma is classified as nottingham grade i and formed a \unk{} mass in the \unk{} region of the breast as well as smaller masses in other regions .} {\color{wSc}\textbf{\scalebox{.85}{3}}\,the smaller masses are associated with extensive lobular carcinoma in situ .} \textcolor{wMuted}{\dots{} (6 more sentences)}}\hfill
\wpanel{3.9}{2.4}{ESP plan ($L{=}164$) $+$ CSP}{{\color{wSa}\textbf{1}\,procedure \textcolor{wKn}{[invasive lobular carcinoma]}}\\{\color{wSb}\textbf{2}\,grade \textcolor{wKn}{[grade 2]}}\\{\color{wSc}\textbf{3}\,in-situ component}\\4\,benign changes \textcolor{wKn}{[fibrocystic change, microcalcifications]}\\5\,other tissue\\6\,margins\\7\,axillary nodes \textcolor{wKn}{[nodal metastasis]}\\8\,vascular invasion \textcolor{wKn}{[lymphovascular invasion]}\\9\,other tissue}\hfill
\wpanel{5.55}{2.4}{Reference}{pathological summary the patients diagnosis is \uline{invasive lobular carcinoma} of the right breast with a tumor size of approximately \unk{} . the tumor is present at the \unk{} lesion site and is classified as \uline{invasive lobular carcinoma} with a \unk{} of classical and focal pleomorphic types . the nuclear grade is \unk{} and there is no in-situ component . lymph node sampling showed positive results with extranodal involvement . the margins for the invasive lobular carcinoma are negative . there is focal suspicion of lymphovascular invasion . no microcalcifications are identified and the skin is negative for tumor involvement . \textcolor{wMuted}{\dots}}\hfill
\wpanel{3.55}{2.4}{\method}{{\color{wSa}\textbf{\scalebox{.85}{1}}\,the pathological slide indicates that the patient underwent a mastectomy for a \uwave{left} breast cancer .} {\color{wSb}\textbf{\scalebox{.85}{2}}\,the tumor is \uline{invasive lobular carcinoma} with a nottingham grade of the tumor is classified as nottingham grade 2 and there is \uwave{no angiolymphatic invasion} .} {\color{wSc}\textbf{\scalebox{.85}{3}}\,the margins are negative for tumor .} \textcolor{wMuted}{\dots}}
\end{minipage}}
\caption{\textbf{\method at work on a test slide} (upper-quartile case of per-report BLEU-4, selected by rule).
$\hat{\mathbf{I}}$: attention of the \ske to every tile; red box: the $16{\times}16$-tile window of highest attention,
magnified. $\hat{\mathbf{T}}$: the five findings with the highest attention in red. CCD: decided claims with calibrated
posteriors. ESP: the most image-similar training report and its plan, with the claims placed by CSP (red). Colours link
planned and written sentences; \uline{underlined}/\uwave{wavy}: as in/opposite to the reference.}
\label{fig:qualitative}
\end{figure*}

\subsection{Comparison with State-of-the-Art Methods}
\label{sec:sota}

\noindent\textbf{Under a common protocol.} As shown in Table~\ref{tab:main}, \method and its EMA variant attain the
highest BLEU-2, BLEU-3 and BLEU-4 under the protocol, and \method's BLEU-4 is significantly higher than that of every
re-trained generator, every slide-level LLM and every retrieval control. Relative to \methodbase, the same network
trained without the decisions, it lengthens the reports from 109 to 143 tokens and raises BLEU-4 from 0.111 to 0.1425
($p<0.001$). The hospital consensus, which generates nothing, reaches the highest BLEU-1 and METEOR: as
Sec.~\ref{sec:analysis} predicts, recall-oriented metrics reward a hospital's reporting format.

\noindent\textbf{Against slide-level LLMs.} Re-trained on our 792 training reports, the slide-level LLMs, with
23 to 217$\times$ the parameters of \method, reach BLEU-4 of 0.078 to 0.082, below MI-Gen and every retrieval control, and
agree with the clinical record significantly less often (macro-F1 0.175 to 0.265 vs.\ 0.328). They under-generate
(98 to 110 tokens) and state frequent findings (\supp{app:llm}). The LLM without visual input writes one report for all
patients and scores as well (BLEU-4 0.079): fine-tuned on 792 reports, the LLMs learn a typical report rather than a
mapping from the slide.

\noindent\textbf{Against published results.} Published numbers are single runs on a slightly different test split
(\supp{app:ceiling}). \method is ahead of every published method
in BLEU-2 and BLEU-3 in all seeds (BLEU-3: $+0.011$ over RANGER, $p=0.02$), on par with RANGER in BLEU-1, BLEU-4
(0.1425 vs.\ 0.1435) and ROUGE-L, and behind it in METEOR ($-0.015$). \method-EMA, selected on validation by a
pre-registered rule (\supp{app:ema}), reaches BLEU-2/3/4 of 0.315/0.217/0.1445, the highest mean values of all methods.

\noindent\textbf{Is the gain only length?} Largely, for BLEU: the length control recovers most of it (0.1378; \method
$+0.005$, n.s.). The decisions change what fills the length: relative to the length control, \method has higher BLEU-3,
ROUGE-L ($+0.013$) and keyword overlap and a lower hallucination rate (0.439 vs.\ 0.476) and clinical discordance (0.272
vs.\ 0.298), all significant (\supp{app:modern}), but it contradicts the clinical record more often (0.232 vs.\ 0.189;
Sec.~\ref{sec:clinical}). Without a decoder, the decisions raise the finding F1 of the retrieved report from 0.39 to
0.50; 42 more tokens leave that of \methodbase at 0.44 (\supp{app:content}).

\noindent\textbf{Generalisation to lung adenocarcinoma.}\label{sec:luad} With all settings of BRCA, \method raises BLEU-4 on
PathText-LUAD (466 patients, 44 for test, from 33 hospitals disjoint from those of BRCA) from 0.097 (\methodbase) to 0.128 ($+0.031$,
95\% CI $[0.022, 0.038]$), leading in BLEU-1 to BLEU-4 and ROUGE-L with unchanged finding F1 (\supp{app:luad}).

\subsection{Clinical Fidelity and Explicit Abstention}
\label{sec:clinical}

The reference reports reach a macro-F1 of 0.515 and a contradiction rate of 0.052, the ceiling of this evaluation.
\method's reports state the seven variables correctly significantly more often than \methodbase
(0.328 vs.\ 0.293, $p=0.009$), MI-Gen and HistGen (0.259, 0.201; $p<0.001$) and slide-level LLMs, and as often as the
length control (0.330), the LLM given \method's decisions (0.324) and the retrieval controls. Per variable (\supp{app:content}), the
decisions improve histology, which the slide shows: applied by rule to the retrieved report, they raise its macro-F1
from 0.35 to 0.49 (\method: 0.52). \method's gain over \methodbase lies in variables H\&E does not determine, stated
more often (65 vs.\ 48\%) at the same accuracy.

\noindent\textbf{Overlap optimisation adds contradictions.} Even a real report of a similar case contradicts the record
in 16 to 27\% of cases. \method's increase over
\methodce (0.232 vs.\ 0.185) arises in \sast ($+0.047$, 95\% CI $[0.021, 0.072]$), which teaches the
decoder to state what reports usually contain: lymph-node status in 95\% of reports (references: 80\%) and receptor
status in 67\% (35 to 39\%), although the slide predicts nodal status only weakly (AUC 0.69) and receptors are measured by
immunohistochemistry (\supp{app:clinical}).

\noindent\textbf{Explicit abstention reverses it.} Withholding receptor status (Sec.~\ref{sec:abstain}) lowers the
contradiction rate from 0.232 to 0.130 (95\% CI of the difference $[-0.121,-0.082]$), below \methodbase (0.169;
$[-0.062,-0.015]$; 23\% fewer) and the length control (0.189), though not below MI-Gen (0.108), which rarely states
receptors (\supp{app:clinical}), while BLEU-4 (0.1265), METEOR and ROUGE-L stay significantly above
those of \methodbase; the price is the correct receptor statements (macro-F1 0.222). Even without
receptor types in its plan, the decoder writes about one receptor sentence per report, which step (iii) withholds
(\supp{app:abstain}).

\subsection{Ablation Study}
\label{sec:ablation}

\noindent\textbf{Effect of the \psd.} The \psd carries the gain in overlap (BLEU-4 0.111 $\rightarrow$ 0.142; Table~\ref{tab:main}) and raises the
clinical macro-F1 (0.293 $\rightarrow$ 0.314) without adding contradictions.

\noindent\textbf{Effect of the \ccd and the \csp.} They change which content fills the structure: over the \psd alone,
they raise CRQS (0.244 $\rightarrow$ 0.280, $p=0.018$) and the F1 of the stated findings (0.424 $\rightarrow$ 0.458,
$p=0.003$; \supp{app:modern}), with BLEU-4 unchanged ($-0.003$, n.s.). Notably,
placement is essential: claims given only as a set (\ccd without \csp) lower BLEU-4 to 0.135 and the macro-F1 to 0.289.

\noindent\textbf{Effect of \sast.} \sast adds 0.012 ROUGE-L ($p<0.001$), 0.004 BLEU-4 ($p=0.06$) and
contradictions (Sec.~\ref{sec:clinical}).

\noindent\textbf{Decisions for a 7B LLM decoder.} Given the decisions in its prompt, TITAN + LLM states the decided findings (finding macro-F1 0.13 $\rightarrow$ 0.32), nearly
doubles its agreement with the clinical record (0.175 $\rightarrow$ 0.324, on par with \method), and its BLEU-1, BLEU-4
and METEOR rise significantly (\supp{app:llm}). The decisions thus
transfer to a pre-trained decoder. Given the five most similar reports
instead, its findings and histology are worse (macro-F1 0.27 and 0.45 vs.\ 0.32 and 0.52) at similar clinical F1 and
higher ROUGE-L; both trail \method in overlap (\supp{app:content}).

\noindent\textbf{Qualitative results.}\label{sec:qualitative} In Fig.~\ref{fig:qualitative}, the \ske attends almost uniformly to the 20 findings, so the findings
come from the \ccd: six claims, led by the reference's histology, placed by the \csp
into the \psd plan of a similar lobular report. The \pcd follows plan and histology but names the wrong side and denies a
claimed vascular invasion (\supp{app:claims}).

\section{Conclusion}
\label{sec:conclusion}

We presented \method, a decide-then-decode generator for whole-slide pathology reports that decides the structure,
the findings and their placement before decoding. With 35M parameters, it outperforms re-trained slide-level LLMs,
matches the strongest published results, transfers to lung cancer from unseen hospitals and can withhold what H\&E
cannot determine. We believe that explicit report-level decisions are a sound basis for faithful report generation.

\noindent\textbf{Limitations and future work.} While \method generalises to an unseen organ and unseen hospitals,
several limitations remain. First, we evaluate two organs of one benchmark family with small test sets (92 and 44
patients). Second, RANGER and BiGen could not be re-trained. Pathologist-verified findings, larger
multi-organ cohorts and decision-conditioned LLMs are worthy future directions.

\section*{LLM Usage Statement}
Claude (Anthropic), accessed through Claude Code, was used to assist with coding,
analysis, and manuscript preparation, including \LaTeX{} compilation and dependency installation. It supported debugging
evaluation and analysis scripts, suggesting robustness checks, editing text and figures, and reviewing related
literature, equations, and references. The authors designed the study, reviewed and executed all code, conducted the
experiments, and independently verified all results, claims, and citations.

{
    \par\small\raggedright
    \bibliographystyle{ieeenat_fullname}
    \bibliography{refs}
}

\clearpage
\onecolumn
\setcounter{table}{0}
\setcounter{figure}{0}
\setcounter{equation}{0}
\setcounter{algorithm}{0}
\renewcommand{\thetable}{S\arabic{table}}
\renewcommand{\thefigure}{S\arabic{figure}}
\renewcommand{\theequation}{S\arabic{equation}}
\renewcommand{\thealgorithm}{S\arabic{algorithm}}
\renewcommand{\theHtable}{S\arabic{table}}
\renewcommand{\theHfigure}{S\arabic{figure}}
\renewcommand{\theHequation}{S\arabic{equation}}
\providecommand{\theHalgorithm}{}
\renewcommand{\theHalgorithm}{S\arabic{algorithm}}

\begin{center}
{\Large\bf \textsc{Clasp}: Deciding Structure and Claims Before Decoding\\for Whole-Slide Pathology Report Generation\par}
\vspace{0.6em}
{\large Supplementary Material\par}
\vspace{1.2em}
\end{center}

\noindent Numbered sections, equations, tables and figures (\eg Sec.~\ref{sec:esp}, Eq.~\eqref{eq:esp},
Table~\ref{tab:main}) refer to the main paper; lettered sections and S-numbered tables and figures refer to this
supplement. Component names are those of the main paper: Slide Knowledge Encoder (\ske), Plan-Conditioned Decoder (\pcd),
Exemplar Structure Prior (\psd), Calibrated Claim Decision (\ccd), Claim-to-Sentence Placement (\csp) and
Structure-Anchored Sequence Training (\sast).

\appendix
\section{Benchmark analysis}
\label{app:benchmark}

\noindent\textbf{Model-free protocol.} Retrieval ranks the 792 training cases by the cosine similarity of the mean UNI2-h
tile embedding. A \emph{consensus} selects, among a candidate set, the report with the highest mean sentence BLEU-4 to the
other candidates~\cite{devlin2015nn}. The hospital is the tissue source site (TSS) of the TCGA case identifier; it is used here as an
oracle, never by a model. Since nothing is trained or selected, validation and test are both reported. The properties
of Sec.~\ref{sec:analysis} hold on both splits; the main paper quotes test values because they can be compared with
published scores.

\noindent\textbf{The hospital is visible in the slide.} The top-1 retrieved report comes from the query's hospital for
91.8\%/82.6\% of validation/test cases (chance 6.7\%); a linear probe on the mean embedding predicts the hospital with
91.8\%/92.4\% accuracy (majority class 21.2\%/15.2\%). Same-hospital report pairs have 2.4 to 2.6$\times$ the sentence BLEU-4
of other pairs (0.076/0.080 vs.\ 0.031/0.030).

\begin{table}[!htb]
\caption{BLEU-4 of report \emph{selection} without learning. ``True findings'' are the 20 silver findings of the query;
``oracle'' rows pick the candidate with the best BLEU-4 against the reference.}
\label{tab:selection}
\centering\footnotesize
\setlength{\tabcolsep}{4pt}
\begin{tabular}{@{}lcc@{}}
\toprule
Selection rule & Val & Test\\
\midrule
Top-1 image-similar report & 0.101 & 0.107\\
Consensus of 30 image-similar & 0.117 & 0.120\\
Consensus of 30 with most similar true findings & 0.096 & 0.113\\
Consensus of 30 random, same hospital (draw 1) & 0.129 & 0.135\\
Consensus of 30 random, same hospital (draw 2) & 0.120 & 0.140\\
Same hospital, 5 best-matching true findings & 0.134 & 0.127\\
Same hospital, 10 best-matching true findings & 0.128 & 0.141\\
Same hospital, 30 best-matching true findings & 0.123 & 0.136\\
\midrule
Best of 5 image-similar (oracle) & 0.142 & 0.152\\
Best of 10 image-similar (oracle) & 0.156 & 0.165\\
Best of 30 image-similar (oracle) & 0.175 & 0.184\\
\bottomrule
\end{tabular}
\end{table}

\begin{figure}[!htb]
\centering
\includegraphics[width=0.5\linewidth, trim=292 0 0 0, clip]{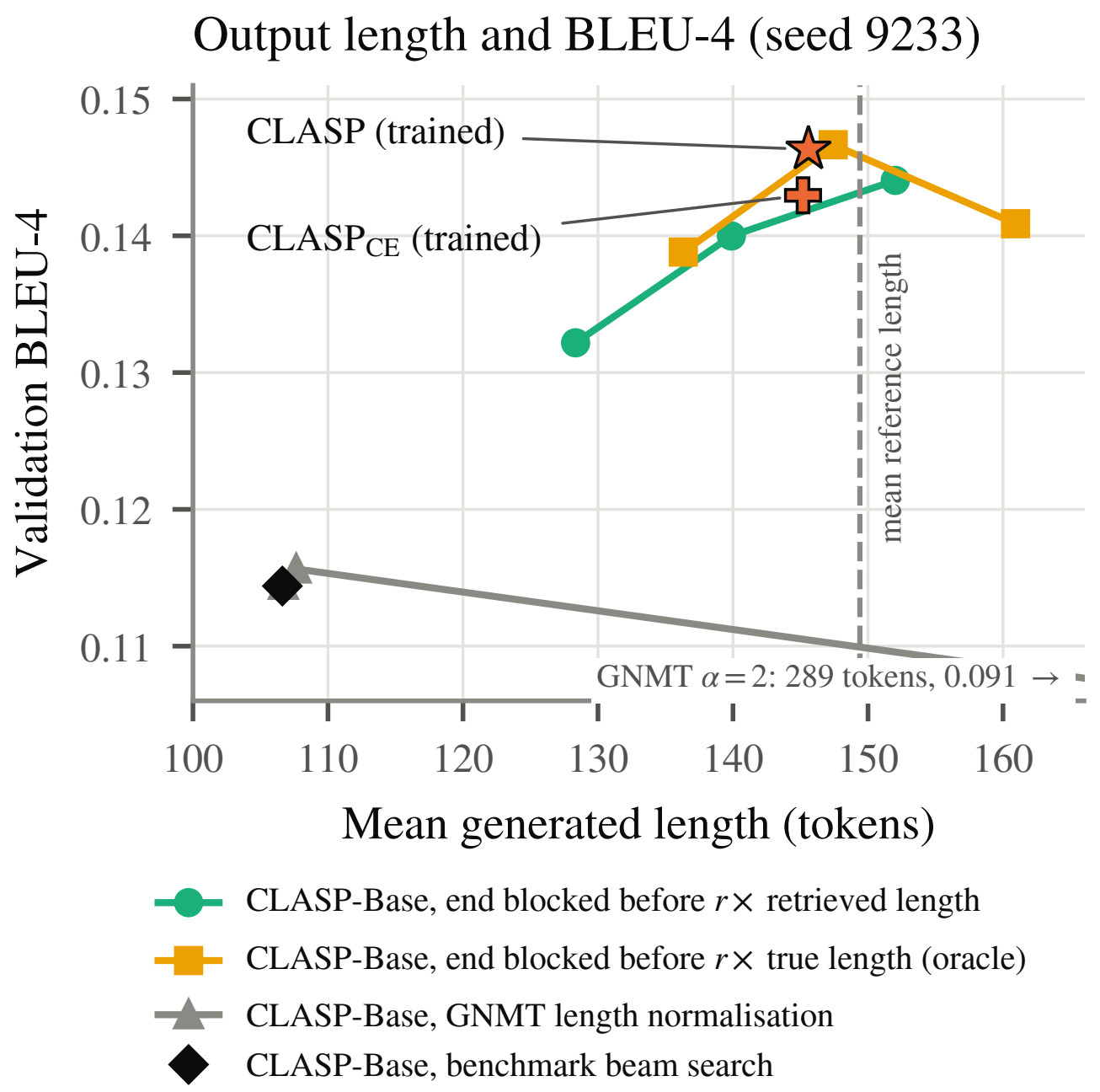}
\caption{\textbf{Output length and validation BLEU-4} (seed 9233): one trained \methodbase decoded with different length
behaviour (lines), and models trained with the \psd (markers). A length taken from retrieved reports recovers most of
what the true reference length gives; length normalisation does not.}
\label{fig:lengthcurve}
\end{figure}

\noindent\textbf{Details of Sec.~\ref{sec:analysis}.} \emph{Hospital.} Restricting the choice to the query's hospital and
ranking its reports by true findings changes BLEU-4 by $-0.012$ to $+0.014$ relative to 30 random reports of that hospital
($k\in\{5,10,30\}$ best-matching reports, validation and test), the same order as the difference between two independent
random draws (0.010 on validation, 0.005 on test; Table~\ref{tab:selection}). \emph{Retrieved content.} Among the
finding statements of the 8 most image-similar training reports, 38.7\% hold for the query (validation: 37.8\%), and
these reports have the right length (test: 144 vs.\ 149 tokens; validation: 147 vs.\ 149); sentences affirming a
finding the query lacks share 1.6\% of their 4-grams with the reference, against 4.6\% for sentences affirming findings
the query has. \emph{Length} (Fig.~\ref{fig:lengthcurve}). Length normalisation of finished beams~\cite{wu2016gnmt} either barely lengthens the outputs
of \methodbase ($\alpha{=}1$: 108 tokens, validation BLEU-4 0.116) or, when strong enough to lengthen them, lets precision
collapse ($\alpha{=}2$: 289 tokens, 0.091). On validation, the length control lowers CRQS from 0.248 to 0.216 and raises
the hallucination rate from 0.476 to 0.507; on test, its CRQS is 0.260 against 0.266 for \methodbase and its
hallucination rate 0.476 against 0.456 (Tables~\ref{tab:main} and~\ref{tab:modern}). \emph{Findings.} \methodbase's
stated findings reach micro-F1 0.410/0.432 (validation/test) and writing the four most frequent training findings
0.437/0.403 (details below).

\noindent\textbf{Where n-gram precision comes from.} Table~\ref{tab:position} splits the clipped 4-gram precision of
validation outputs by the position at which each 4-gram ends (clipping is applied over the whole report, in position
order). All generators are most precise at the beginning of a report. Tokens written after position 100 match the
reference at 0.010 to 0.031, the level of a retrieved report of another patient (0.014 to 0.023). Length control therefore
removes the brevity penalty with tokens of low precision. Structure conditioning splits more sentences into fragments of at
most three words (2.5 to 5.0\% vs.\ 0.5 to 1.1\%) and produces more word 4-grams that never occur in training reports.

\begin{table}[!htb]
\caption{Validation outputs: clipped 4-gram precision by the token position at which the 4-gram ends, share of sentences
with at most three words (enumeration markers excluded), and share of 4-grams absent from all training reports.
Checkpoints are the validation-selected ones. ``n/a'': fewer than 1\% of the system's 4-grams.}
\label{tab:position}
\centering\footnotesize
\setlength{\tabcolsep}{2.6pt}
\begin{tabular}{@{}lrccccrr@{}}
\toprule
 & Len. & 1 to 50 & 51 to 100 & 101 to 150 & 151+ & Frag.\,\% & Novel\,\% \\
\midrule
References & 149 & n/a & n/a & n/a & n/a & 1.1 & 51.1\\
\methodbase & 107 & 0.069 & 0.046 & 0.031 & n/a & 0.5 & 15.1\\
\methodbase + length control & 152 & 0.069 & 0.045 & 0.025 & 0.015 & 0.7 & 21.0\\
\psd, full & 145 & 0.068 & 0.042 & 0.024 & 0.010 & 3.8 & 25.3\\
\psd, budget only & 146 & 0.067 & 0.042 & 0.019 & 0.016 & 3.2 & 24.4\\
\psd, plan only & 143 & 0.070 & 0.040 & 0.016 & 0.014 & 5.0 & 24.4\\
\psd, full, medoid structure & 146 & 0.073 & 0.040 & 0.024 & 0.016 & 2.5 & 26.7\\
\methodce & 145 & 0.061 & 0.048 & 0.030 & 0.012 & 4.9 & 24.8\\
Consensus of 30 image-similar & 151 & 0.049 & 0.028 & 0.021 & 0.023 & 1.0 & 0.0\\
Top-1 image-similar & 151 & 0.038 & 0.021 & 0.018 & 0.014 & 1.1 & 0.0\\
\bottomrule
\end{tabular}
\end{table}

\noindent\textbf{Findings written vs.\ findings the slide supports.} The clinical-efficacy extractor applied to the generated
reports gives \methodbase a micro-F1 of 0.410/0.432 (validation/test). Writing the $k$ most frequent training findings into every
report gives 0.262/0.276 ($k{=}1$), 0.333/0.346 ($k{=}2$), 0.405/0.382 ($k{=}3$) and 0.437/0.403 ($k{=}4$: invasive ductal
carcinoma, DCIS, lymph-node metastasis, grade 2). A linear probe on the mean UNI2-h embedding with per-finding F1-optimal thresholds (five-fold out-of-fold predictions on
the training split) reaches 0.471/0.458 while affirming 6.9/6.7 findings per
report (true: 3.9/3.7). Its slide-level topic classifier with the decision rule of Sec.~\ref{sec:claims} (one global threshold
on calibrated posteriors, chosen on validation) reaches 0.518/0.499 with 5.7/5.6 claims per report (seed 9233;
0.510/0.491 and 0.508/0.510 for seeds 1 and 2).

\section{Headroom and comparison with published numbers}
\label{app:ceiling}

\noindent\textbf{Headroom (Table~\ref{tab:ceiling}).} Giving \methodce the reference's own sentence types and length
raises its validation BLEU-4 from 0.142 to 0.151; either part alone adds at most 0.004, and adding the reference's
findings per sentence adds 0.001 more. Neither part is predictable from the slide: length predictors (ridge regression on
the slide embedding or posteriors, the median length of the predicted hospital) miss the reference length by 47 to 54
tokens on validation, no better than the \psd budget (50 tokens). The remaining BLEU thus lies largely in quantities that a
slide does not determine. This analysis uses validation decodes of the three \methodce checkpoints, was made after the
test evaluations and changed no model. A uniform scaling of the \psd budget (factors 1.0 to 1.2), selected on validation by
a rule fixed in advance, kept the factor 1.0 for every model, so no test decoding was run.

\begin{table}[!htb]
\caption{\textbf{Headroom analysis.} Top: validation BLEU-4 of the three \methodce seeds (mean \pmstd{std}) decoded with
parts of the test-time structure replaced by the reference's; ``best of $K$'': among the plans of the $K$ most
image-similar training reports, the one closest to the reference plan. Middle: mean absolute error (tokens) of length
predictors, validation / test (reference length $149\pm71$ / $149\pm73$). Bottom: share of references whose first four
words (the writing style) equal those of the named report, validation / test. $^{\circ}$: uses the reference or the
hospital, not available at test time. Details: Sec.~\ref{app:ceiling}.}
\label{tab:ceiling}
\centering\footnotesize
\setlength{\tabcolsep}{3pt}
\begin{tabular}{@{}llc@{}}
\toprule
Sentence types & Length & Val BLEU-4\\
\midrule
prior (\psd) & prior (\psd) & 0.142\pmstd{0.002}\\
best of 10 retrieved$^{\circ}$ & prior & 0.141\pmstd{0.002}\\
best of 30 retrieved$^{\circ}$ & prior & 0.143\pmstd{0.003}\\
reference$^{\circ}$ & prior & 0.144\pmstd{0.004}\\
prior & reference$^{\circ}$ & 0.146\pmstd{0.003}\\
reference$^{\circ}$ & reference$^{\circ}$ & 0.151\pmstd{0.002}\\
\quad + reference findings per sentence$^{\circ}$ & reference$^{\circ}$ & 0.152\pmstd{0.002}\\
\midrule
\multicolumn{2}{@{}l}{Length predictor} & MAE\\
\midrule
\multicolumn{2}{@{}l}{constant (training mean)} & 56.2 / 49.4\\
\multicolumn{2}{@{}l}{mean of the 5 most similar reports (\psd)} & 50.0 / 46.7\\
\multicolumn{2}{@{}l}{ridge regression on the slide embedding} & 46.9 / 48.0\\
\multicolumn{2}{@{}l}{median length of the true hospital$^{\circ}$} & 46.1 / 44.4\\
\multicolumn{2}{@{}l}{ridge regression on the true findings$^{\circ}$} & 47.2 / 39.2\\
\multicolumn{2}{@{}l}{best of the 10 most similar reports$^{\circ}$} & 12.5 / 14.8\\
\midrule
\multicolumn{2}{@{}l}{Opening words equal to those of} & share\\
\midrule
\multicolumn{2}{@{}l}{the most frequent training opening} & 0.19 / 0.20\\
\multicolumn{2}{@{}l}{the most image-similar training report} & 0.08 / 0.08\\
\multicolumn{2}{@{}l}{a random report of the same hospital$^{\circ}$} & 0.07 / 0.08\\
\multicolumn{2}{@{}l}{a random training report} & 0.07 / 0.07\\
\bottomrule
\end{tabular}
\end{table}

\begin{figure}[!htb]
\centering
\includegraphics[width=0.55\linewidth]{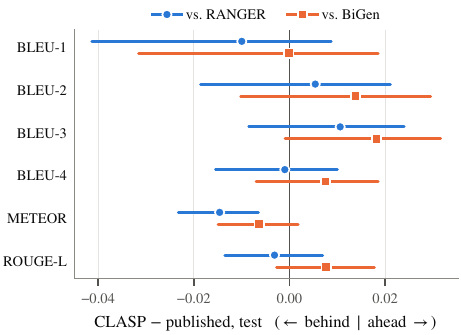}
\caption{\textbf{\method against the strongest published results.} Difference between \method's seed-averaged test
score and the quoted single runs of RANGER~\cite{chen2026ranger} and BiGen~\cite{zhang2025bigen}; bars: 95\%
patient-bootstrap interval of \method's score. Exact values: Table~\ref{tab:ranger}.}
\label{fig:published}
\end{figure}

\noindent\textbf{Comparison with published numbers (Table~\ref{tab:ranger}).} A patient bootstrap resamples the 92 test
patients (10{,}000 times, seed 0), and in every resample BLEU-4 is recomputed from summed $n$-gram statistics and averaged
over the three seeds. The $t$-test uses the three per-seed values only (seed variation). Neither accounts for the
differences in test set (93 vs.\ 92 patients) and features, or for the unknown seed variation of the published runs.
Table~\ref{tab:ranger} gives all six published metrics for \method (the values plotted in Fig.~\ref{fig:published}).

\begin{table}[!htb]
\caption{\textbf{\method against the best published run} (RANGER~\cite{chen2026ranger}, quoted). Mean \pmstd{std} over three seeds; 95\% interval of the seed-averaged test metric from a bootstrap over the 92 test patients; number of seeds at or above the quoted value; two-sided one-sample $t$-test of the three seeds against it. The intervals cover only our own sampling uncertainty; the published run has its own, unreported one.}
\label{tab:ranger}
\centering\footnotesize
\setlength{\tabcolsep}{2.6pt}
\begin{tabular}{@{}lccccc@{}}
\toprule
Metric & RANGER & \method & 95\% interval & Seeds $\ge$ & $p$\\
\midrule
BLEU-1 & 0.4598 & 0.450\pmstd{0.007} & [0.418, 0.469] & 0/3 & 0.14\\
BLEU-2 & 0.3044 & 0.310\pmstd{0.003} & [0.286, 0.325] & 3/3 & 0.08\\
BLEU-3 & 0.2036 & 0.214\pmstd{0.003} & [0.195, 0.228] & 3/3 & 0.02\\
BLEU-4 & 0.1435 & 0.1425\pmstd{0.0026} & [0.1280, 0.1535] & 1/3 & 0.57\\
METEOR & 0.1883 & 0.174\pmstd{0.002} & [0.165, 0.182] & 0/3 & 0.00\\
ROUGE-L & 0.3038 & 0.301\pmstd{0.004} & [0.290, 0.311] & 1/3 & 0.27\\
\bottomrule
\end{tabular}
\end{table}

 All three \method seeds exceed BiGen's quoted 0.135 ($t$-test $p=0.04$, seed variation only), but
BiGen's value lies inside the patient-bootstrap interval.

\section{Independent clinical evaluation: details}
\label{app:clinical}

\noindent\textbf{Ground truth.} The TCGA-BRCA clinical supplement, obtained from the NCI Genomic Data Commons, records
per patient the variables abstracted by the TCGA Biospecimen Core Resource~\cite{liu2018tcgacdr}. It is used for evaluation only: no model, threshold
or checkpoint depends on it. Patients whose value is missing, not evaluated or indeterminate are excluded for that
variable (Table~\ref{tab:clinvars}).

\begin{table}[!htb]
\caption{Variables of the independent clinical evaluation: truth classes (from the TCGA record), number of test patients
with a truth value, and the reader's options.}
\label{tab:clinvars}
\centering\footnotesize
\setlength{\tabcolsep}{4pt}
\begin{tabular}{@{}llcl@{}}
\toprule
Variable & Truth classes (TCGA record) & $n$ test & Reader options\\
\midrule
Histology & IDC / ILC / other incl.\ mixed & 92 & IDC, ILC, other or mixed, not stated\\
ER & positive / negative & 88 & positive, negative, not stated\\
PR & positive / negative & 88 & positive, negative, not stated\\
HER2 & positive / negative / equivocal & 75 & positive, negative, equivocal, not stated\\
Lymph nodes & pN0$^{*}$ / pN1+ incl.\ N1mi & 92 & metastasis present, nodes negative, not stated\\
Margin & positive / close / negative & 90 & positive, close, negative, not stated\\
Tumour size & T1 / T2 / T3; T4 excluded & 88 & $\le$2\,cm, 2 to 5\,cm, $>$5\,cm, not stated\\
\bottomrule
\end{tabular}
\end{table}

\noindent\textbf{Reader.} Llama-3.1-8B-Instruct~\cite{grattafiori2024llama3} in bfloat16. Every question is posed as a
chat whose system message instructs the model to extract facts from pathology reports exactly as written; the user
message contains the report, the question and the lettered options, and asks for the letter of one option, answered
only from what the report states. The answer is the option letter with the highest next-token logit, so reading is
deterministic, and each report is read once per variable. The reader receives only the report text.

\noindent\textbf{Reader reliability.} Table~\ref{tab:reader} gives, for the reference reports, how often each variable is
stated and how often the reader's answer agrees with the TCGA record when it is. Disagreements combine reader errors and
genuine differences between the free-text report and the abstracted record; they bound the precision any system can
reach. The references themselves reach a mean macro-F1 of 0.515 and a contradiction rate of 0.052 on test.

\begin{table}[!htb]
\caption{Reference reports read by the reader: share of patients whose report states the variable, and agreement with the
TCGA record when stated (validation / test).}
\label{tab:reader}
\centering\footnotesize
\setlength{\tabcolsep}{4pt}
\begin{tabular}{@{}lccccccc@{}}
\toprule
 & Histology & ER & PR & HER2 & Nodes & Margin & Size\\
\midrule
Stated & 0.73 / 0.74 & 0.40 / 0.39 & 0.37 / 0.35 & 0.34 / 0.36 & 0.80 / 0.80 & 0.56 / 0.50 & 0.57 / 0.58\\
Agreement when stated & 0.95 / 0.90 & 0.94 / 0.94 & 0.93 / 0.94 & 0.80 / 0.85 & 0.94 / 0.96 & 0.76 / 0.87 & 0.83 / 0.84\\
\bottomrule
\end{tabular}
\end{table}

\noindent\textbf{What the slide can support.} A logistic regression on the mean tile embedding, trained on the training
patients' records, separates the variables with ROC-AUC (validation / test): invasive lobular vs.\ other histology
0.92 / 0.93, ER-negative 0.89 / 0.85, PR-negative 0.73 / 0.80, tumour $\le$2\,cm 0.75 / 0.87, HER2-positive 0.64 / 0.69,
lymph-node metastasis 0.66 / 0.69, positive margin 0.65 / 0.65. Receptor status is measured by immunohistochemistry and
lymph nodes and margins are separate specimens or locations, so statements about them are largely not supported by the
diagnostic H\&E slide; a clinically safe generator should state them rarely, which the contradiction rate measures.

\noindent\textbf{Results per variable (Table~\ref{tab:clinvar}).} \method agrees with the record better than \methodbase on
the three receptors and lymph nodes, equally well on the margin, and worse on histology and tumour size (size stated in
23\% of its reports, against 43\%). Its additional contradictions over
\methodce concentrate on estrogen and progesterone receptor status (0.28 $\rightarrow$ 0.34 and 0.26 $\rightarrow$ 0.36)
and lymph-node status (0.35 $\rightarrow$ 0.46): \sast raises the share of reports that state these variables
(receptors 0.48 to 0.50 $\rightarrow$ 0.67, nodes 0.88 $\rightarrow$ 0.95, references 0.35 to 0.39 and 0.80), i.e., it
learns to state what reports usually contain, including where the slide carries little evidence. The hospital consensus
shows the same pattern more strongly. MI-Gen contradicts least because it rarely states the receptors (19 to 22\% of reports).
The side of the breast shows the same mechanism outside the seven variables: no slide shows which breast it comes from,
yet \method names \emph{both} sides in 67 of the 92 test reports (\methodbase: 11; references: 12), because a report that
names both always matches the reference's side.

\begin{table}[!htb]
\caption{\textbf{Independent clinical evaluation per variable} (test; macro-F1 over the classes of the TCGA record, ``not stated'' counted as a miss / contradiction rate, the share of patients whose report states a value that contradicts the record). Mean over seeds where several were run.}
\label{tab:clinvar}
\centering\footnotesize
\setlength{\tabcolsep}{3pt}
\begin{tabular}{@{}lccccccc@{}}
\toprule
Reports & Histology & ER & PR & HER2 & Nodes & Margin & pT\\
\midrule
Reference reports & 0.55 / 0.08 & 0.51 / 0.02 & 0.48 / 0.02 & 0.35 / 0.05 & 0.85 / 0.03 & 0.27 / 0.07 & 0.59 / 0.09\\
\midrule
MI-Gen~\cite{chen2024wsicaption} & 0.55 / 0.07 & 0.13 / 0.07 & 0.21 / 0.07 & 0.10 / 0.07 & 0.44 / 0.28 & 0.21 / 0.06 & 0.17 / 0.15\\
HistGen~\cite{guo2024histgen} & 0.44 / 0.16 & 0.04 / 0.22 & 0.10 / 0.18 & 0.12 / 0.08 & 0.34 / 0.24 & 0.20 / 0.02 & 0.17 / 0.23\\
Hospital consensus & 0.44 / 0.20 & 0.35 / 0.19 & 0.29 / 0.32 & 0.31 / 0.28 & 0.46 / 0.36 & 0.27 / 0.12 & 0.23 / 0.40\\
\methodbase & 0.58 / 0.11 & 0.23 / 0.18 & 0.26 / 0.18 & 0.14 / 0.10 & 0.38 / 0.34 & 0.27 / 0.08 & 0.18 / 0.20\\
\quad + \psd + \ccd + \csp = \methodce & 0.52 / 0.11 & 0.27 / 0.28 & 0.32 / 0.26 & 0.14 / 0.10 & 0.56 / 0.35 & 0.23 / 0.07 & 0.17 / 0.12\\
\quad + \sast = \textbf{\method (ours)} & 0.52 / 0.13 & 0.36 / 0.34 & 0.37 / 0.36 & 0.16 / 0.10 & 0.48 / 0.46 & 0.28 / 0.10 & 0.12 / 0.13\\
\bottomrule
\end{tabular}
\end{table}

\section{Explicit abstention}
\label{app:abstain}

\noindent\textbf{Decision.} The variables to withhold are fixed by clinical definition, before any decoding and without
data: receptor status (ER, PR, HER2), which is measured by immunohistochemistry or in-situ hybridisation rather than on
the H\&E slide, and, as a second operating point, also lymph-node status, which is assessed on the lymph-node specimens.
No threshold is tuned, the TCGA records are used for evaluation only, and each operating point was decoded once on the
test split for the three seeds of \method; the trained models are unchanged.

\noindent\textbf{Realisation with \method's own components.} (i) \emph{Plan.} The sentence types of a variable are the
types whose training sentences mention it in at least 90\% of cases: two receptor types (98.4\% and 98.6\%) and two
lymph-node types (99.7\% and 98.5\%; the next node type, at 71.6\%, is not withheld). They are removed from the \psd plan
(0.48 planned sentences per report for the receptors, 1.54 with the nodes), and the token budget shrinks by their share of
the planned sentence lengths. (ii) \emph{Claims.} The claims of the withheld findings (estrogen receptor positive;
lymph-node metastasis and extranodal extension) are withdrawn, and \csp places the remaining claims in the reduced plan.
(iii) \emph{Guarantee.} Every generated sentence that the type model assigns to a withheld type is not emitted. With no
variable withheld, this pipeline reproduces the released reports of \method exactly.

\noindent\textbf{Findings.} Removing a type from the plan does not stop the sequence-trained decoder from writing it:
step (iii) withholds 90, 109 and 72 receptor sentences in the three seeds (about one per report), so an explicit
decision has to be enforced, not only conditioned on. Withholding the receptors removes almost all receptor statements
(ER stated in 6\% of reports) and their contradictions (Table~\ref{tab:abstain}). Lymph-node status is also mentioned
within sentences of other types (procedure, other tissue), so withholding its two types leaves it stated in 69\% of
reports and reduces its contradictions only from 0.46 to 0.39, at a large cost in overlap. A complete abstention from a
variable therefore needs a statement-level rather than a sentence-level realisation.

\begin{table}[!htb]
\caption{\textbf{Explicit abstention} (test, mean over three seeds). Len: tokens per report. F1 / Contr.: mean macro-F1
and contradiction rate over the seven TCGA variables. ER, PR, Nodes: share of reports that state the variable /
contradiction rate. Paired differences \method $-$ abstention (receptors), 95\% CI: BLEU-4 $+0.015$ $[0.006, 0.020]$, contradiction $+0.101$ $[0.082, 0.121]$.}
\label{tab:abstain}
\centering\footnotesize
\setlength{\tabcolsep}{4pt}
\begin{tabular}{@{}lccccccccc@{}}
\toprule
 & B-4 & MTR & R-L & Len & F1 & Contr.$\downarrow$ & ER & PR & Nodes\\
\midrule
\methodbase & 0.1106 & 0.150 & 0.287 & 109 & 0.293 & 0.169 & 0.38 / 0.18 & 0.38 / 0.18 & 0.68 / 0.34\\
\methodce & 0.1383 & 0.172 & 0.289 & 143 & 0.313 & 0.185 & 0.48 / 0.28 & 0.50 / 0.26 & 0.88 / 0.35\\
\midrule
\method & 0.1425 & 0.174 & 0.301 & 143 & 0.328 & 0.232 & 0.67 / 0.34 & 0.67 / 0.36 & 0.95 / 0.46\\
\quad abstain: receptors & 0.1265 & 0.162 & 0.298 & 122 & 0.222 & 0.130 & 0.06 / 0.04 & 0.05 / 0.02 & 0.92 / 0.48\\
\quad abstain: receptors, nodes & 0.1022 & 0.143 & 0.294 & 95 & 0.187 & 0.116 & 0.07 / 0.06 & 0.03 / 0.03 & 0.69 / 0.39\\
\bottomrule
\end{tabular}
\end{table}

\section{2025 to 2026 evaluation protocols}
\label{app:modern}

\noindent\textbf{2025 to 2026 protocols (Table~\ref{tab:modern}).} We score every method with the ranking score of the
REG~2025 challenge~\cite{lee2026reg}, PathReportEval's clinical report quality score CRQS and its parts~\cite{singh2026pathreporteval},
BERTScore~\cite{zhang2020bertscore}, and the micro-F1 of the 20 findings of the rule-based extractor (CE-F1). CE-F1 is not
independent of \method: the same extractor defines the training labels of the slide-level model that decides the claims,
so we use it only to compare the components of \method, and the TCGA records (Sec.~\ref{sec:setup}) as the independent
clinical measure. Four observations. (i) \ccd and \csp raise CRQS from 0.244 (\psd) to 0.280 and CE-F1
from 0.424 to 0.458; \sast returns part of this gain (CRQS 0.269, CE-F1 0.439) while lowering the clinical discordance
score (CDS) to the lowest value of all methods (0.272). (ii) BERTScore is lowest for \method (0.117). Conditioning on a
planned sequence of sentences closes sentences early (4.9\% of \methodce's sentences have at most three words, against
0.5\% for \methodbase; Table~\ref{tab:position}); BERTScore penalises such fragments and BLEU does not. (iii) The
embedding-based parts of the REG score (EMB) and BERTScore are highest for the retrieval baselines, which return fluent
human-written text. (iv) The re-trained slide-level LLMs have the lowest CRQS of all methods (0.157 to 0.207), and their
REG score (0.320 to 0.329) is below that of every other learned generator (0.334 to 0.360); given \method's decisions, the
LLM's CRQS rises to 0.248 and its hallucination rate falls from 0.517 to 0.484 (\supp{app:llm}).

\begin{table}[!htb]
\caption{\textbf{2025 to 2026 evaluation protocols} (test; methods run under our protocol). REG: ranking score of the REG~2025 challenge~\cite{lee2026reg} ($0.15\,$(ROUGE-L + BLEU-4 precision) $+\,0.4\,$KEY $+\,0.3\,$EMB; KEY: Jaccard of scispaCy entities; EMB: Llama3-OpenBioLLM-8B embedding similarity). CRQS of PathReportEval~\cite{singh2026pathreporteval} and its parts: clinical field coverage (CFC), key information recall (KIR), hallucination rate (HR) and clinical discordance (CDS), lower is better for the last two. BERTScore: roberta-large, rescaled~\cite{zhang2020bertscore}. CE-F1: micro-F1 of the 20 findings of the rule-based extractor (generated vs.\ reference report; the same extractor defines the training labels). Marks as in Table~\ref{tab:main}.}
\label{tab:modern}
\centering\footnotesize
\setlength{\tabcolsep}{3pt}
\resizebox{\textwidth}{!}{%
\begin{tabular}{@{}lcccccccccc@{}}
\toprule
Method & REG & KEY & EMB & CRQS & CFC & KIR & HR$\downarrow$ & CDS$\downarrow$ & BERTSc. & CE-F1\\
\midrule
MI-Gen~\cite{chen2024wsicaption} & 0.355 & 0.158$^{*}$ & 0.788$^{\dagger}$ & 0.259 & 0.505 & 0.566 & \second{0.416} & \second{0.284} & 0.156$^{\dagger}$ & 0.446\\
HistGen~\cite{guo2024histgen} & 0.334$^{*}$ & 0.134$^{*}$ & 0.757$^{*}$ & 0.236 & 0.482$^{*}$ & 0.555 & \best{0.414} & 0.297 & 0.152$^{\dagger}$ & 0.374$^{*}$\\
HistoGPT-1B~\cite{tran2025histogpt} & 0.329\pmstd{0.012}$^{*}$ & 0.121\pmstd{0.011}$^{*}$ & 0.775\pmstd{0.022} & 0.157\pmstd{0.064}$^{*}$ & 0.468\pmstd{0.047}$^{*}$ & 0.506\pmstd{0.062}$^{*}$ & 0.486\pmstd{0.009}$^{*}$ & 0.340\pmstd{0.016}$^{*}$ & 0.190\pmstd{0.012}$^{\dagger}$ & 0.360\pmstd{0.041}$^{*}$\\
SlideChat arch.~\cite{chen2025slidechat} (7B) & 0.320\pmstd{0.010}$^{*}$ & 0.111\pmstd{0.011}$^{*}$ & 0.780\pmstd{0.028} & 0.185\pmstd{0.073}$^{*}$ & 0.490\pmstd{0.029}$^{*}$ & 0.534\pmstd{0.058}$^{*}$ & 0.493\pmstd{0.048}$^{*}$ & 0.331\pmstd{0.030}$^{*}$ & 0.163\pmstd{0.001}$^{\dagger}$ & 0.321\pmstd{0.011}$^{*}$\\
TITAN~\cite{ding2025titan} + LLM (7B) & 0.324\pmstd{0.006}$^{*}$ & 0.110\pmstd{0.010}$^{*}$ & 0.792\pmstd{0.003}$^{\dagger}$ & 0.207\pmstd{0.030}$^{*}$ & 0.515\pmstd{0.011} & 0.566\pmstd{0.009} & 0.517\pmstd{0.030}$^{*}$ & 0.331\pmstd{0.026}$^{*}$ & 0.168\pmstd{0.002}$^{\dagger}$ & 0.342\pmstd{0.018}$^{*}$\\
\quad + \method decisions as text & 0.332\pmstd{0.003}$^{*}$ & 0.121\pmstd{0.006}$^{*}$ & 0.803\pmstd{0.002}$^{\dagger}$ & 0.248\pmstd{0.034} & \second{0.549}\pmstd{0.027}$^{\dagger}$ & 0.579\pmstd{0.027} & 0.484\pmstd{0.008}$^{*}$ & 0.314\pmstd{0.017}$^{*}$ & 0.169\pmstd{0.004}$^{\dagger}$ & \second{0.457}\pmstd{0.013}\\
LLM without visual input (7B) & 0.324\pmstd{0.006}$^{*}$ & 0.111\pmstd{0.007}$^{*}$ & 0.792\pmstd{0.013}$^{\dagger}$ & 0.188\pmstd{0.046}$^{*}$ & 0.477\pmstd{0.070}$^{*}$ & 0.547\pmstd{0.028}$^{*}$ & 0.490\pmstd{0.046}$^{*}$ & 0.331\pmstd{0.024}$^{*}$ & 0.164\pmstd{0.007}$^{\dagger}$ & 0.320\pmstd{0.029}$^{*}$\\
Most similar training report & 0.348$^{*}$ & 0.143$^{*}$ & 0.808$^{\dagger}$ & 0.232 & 0.524 & 0.560 & 0.468 & 0.313$^{*}$ & 0.192$^{\dagger}$ & 0.435\\
Consensus of 30 similar~\cite{devlin2015nn} & 0.355 & 0.149$^{*}$ & \second{0.818}$^{\dagger}$ & 0.245 & \best{0.556} & 0.561 & 0.478$^{*}$ & 0.310 & \second{0.208}$^{\dagger}$ & 0.418\\
Hospital consensus & \best{0.367}$^{\dagger}$ & 0.157$^{*}$ & \best{0.839}$^{\dagger}$ & 0.229 & 0.539 & 0.535 & 0.487$^{*}$ & 0.296 & \best{0.224}$^{\dagger}$ & 0.446\\
\midrule
\methodbase{} + length control & 0.358\pmstd{0.011} & 0.162\pmstd{0.011}$^{*}$ & 0.810\pmstd{0.013}$^{\dagger}$ & 0.260\pmstd{0.073} & 0.543\pmstd{0.050}$^{\dagger}$ & 0.584\pmstd{0.051} & 0.476\pmstd{0.024}$^{*}$ & 0.298\pmstd{0.027}$^{*}$ & 0.154\pmstd{0.043}$^{\dagger}$ & 0.445\pmstd{0.029}\\
\methodbase & 0.354\pmstd{0.010} & 0.153\pmstd{0.011}$^{*}$ & 0.800\pmstd{0.008}$^{\dagger}$ & 0.266\pmstd{0.071} & 0.532\pmstd{0.056} & 0.580\pmstd{0.052} & 0.456\pmstd{0.021} & 0.286\pmstd{0.020} & 0.177\pmstd{0.040}$^{\dagger}$ & 0.439\pmstd{0.034}\\
\quad + \psd & \second{0.360}\pmstd{0.003}$^{\dagger}$ & 0.165\pmstd{0.002}$^{*}$ & 0.810\pmstd{0.003}$^{\dagger}$ & 0.244\pmstd{0.037} & 0.524\pmstd{0.015} & 0.560\pmstd{0.024} & 0.453\pmstd{0.018} & 0.298\pmstd{0.025}$^{*}$ & 0.159\pmstd{0.011}$^{\dagger}$ & 0.424\pmstd{0.023}\\
\quad + \psd + \ccd & 0.354\pmstd{0.002} & 0.158\pmstd{0.002}$^{*}$ & 0.804\pmstd{0.006}$^{\dagger}$ & 0.245\pmstd{0.023} & 0.527\pmstd{0.027} & 0.576\pmstd{0.020} & 0.469\pmstd{0.020}$^{*}$ & 0.307\pmstd{0.006}$^{*}$ & 0.139\pmstd{0.013}$^{\dagger}$ & 0.455\pmstd{0.011}\\
\quad + \psd + \ccd + \csp = \methodce & 0.355\pmstd{0.009} & 0.161\pmstd{0.008}$^{*}$ & 0.800\pmstd{0.017}$^{\dagger}$ & \best{0.280}\pmstd{0.009} & 0.543\pmstd{0.023}$^{\dagger}$ & \second{0.589}\pmstd{0.018}$^{\dagger}$ & 0.443\pmstd{0.036} & \second{0.284}\pmstd{0.011} & 0.136\pmstd{0.029}$^{\dagger}$ & \best{0.458}\pmstd{0.016}\\
\quad + \sast = \textbf{\method (ours)} & 0.357\pmstd{0.006} & \best{0.176}\pmstd{0.004} & 0.778\pmstd{0.017} & \second{0.269}\pmstd{0.030} & 0.526\pmstd{0.021} & 0.567\pmstd{0.016} & 0.439\pmstd{0.013} & \best{0.272}\pmstd{0.015} & 0.117\pmstd{0.025} & 0.439\pmstd{0.035}\\
\quad \method-EMA (validation-selected) & 0.358\pmstd{0.006} & \second{0.171}\pmstd{0.008}$^{*}$ & 0.791\pmstd{0.020}$^{\dagger}$ & 0.267\pmstd{0.029} & 0.526\pmstd{0.019} & \best{0.590}\pmstd{0.018}$^{\dagger}$ & 0.454\pmstd{0.008} & 0.291\pmstd{0.016}$^{*}$ & 0.133\pmstd{0.036}$^{\dagger}$ & 0.447\pmstd{0.025}\\
\bottomrule
\end{tabular}}
\end{table}

\section{Do the decisions change content or only length?}
\label{app:content}

The length control recovers most of \method's gain in BLEU-4 (Sec.~\ref{sec:sota}), and \method's clinical macro-F1
equals that of the length control and of the retrieval controls (Sec.~\ref{sec:clinical}). Three further checks, fixed
before they were run, test whether the decisions change what a report states and not only how long it is: the decisions
applied by rule to the retrieved report, without any decoder (Table~\ref{tab:contentrule}); results per clinical
variable, split by what the H\&E slide shows (Table~\ref{tab:contentvar}); and the 7B LLM given the retrieved reports as
text instead of the decisions (Table~\ref{tab:llmdecisions}). Scoring and tests are those of every other result.

\noindent\textbf{The decisions without a decoder (Table~\ref{tab:contentrule}).} Per test patient and seed, we take the
\psd's retrieved report (the training report whose plan is the prior plan), delete every sentence that states a finding
the \ccd did not claim, and insert every claimed finding that no kept sentence states: the first sentence of the ten most
image-similar training reports that states it and only claimed findings (else the most frequent such training sentence),
placed after the planned sentence where the \csp places the finding. Nothing is trained or tuned. On the same retrieved
text, the decisions raise the micro-F1 of the stated findings from 0.393 to 0.500 and their macro-F1 from 0.282 to 0.368,
the clinical macro-F1 from 0.306 to 0.335 and CRQS from 0.238 to 0.289, with BLEU-4 unchanged and ROUGE-L and CIDEr
lower. The edits lengthen the reports (139 $\rightarrow$ 167 tokens), but length does not explain the gain: the length
control adds 42 tokens to \methodbase and moves its finding micro-F1 only from 0.439 to 0.445, in the seed whose edits
add 8 tokens the finding micro-F1 still reaches 0.513, and the clinical gain lies in histology, which the slide shows
(below), not in stating more variables. Compared with \method, the rule-based realisation states the decided findings
more completely (finding micro-F1 0.500 vs.\ 0.439) and contradicts the record less often (0.176 vs.\ 0.232) at equal
clinical macro-F1 (0.335 vs.\ 0.328, n.s.), while \method's decoder writes reports much closer to the references
(BLEU-4 $+0.031$, ROUGE-L $+0.027$, CIDEr $+0.066$) with a lower hallucination rate (0.439 vs.\ 0.469). The content is
thus carried by the decisions; the decoder turns them into fluent reports but does not write every decided finding
(Table~\ref{tab:realise}).

\begin{table}[!htb]
\caption{\textbf{\method's decisions applied by rule to the retrieved report} (test; means over three seeds, the
unedited report is deterministic). $p$: one-sided paired bootstrap over test patients (10{,}000 resamples) of the
edited report against the unedited one and against \method, in the direction of the observed difference.}
\label{tab:contentrule}
\centering\footnotesize
\setlength{\tabcolsep}{3pt}
\begin{tabular}{@{}lccccc@{}}
\toprule
Metric & Retrieved, unedited & + edits by rule & $p$ & \method & $p$\\
\midrule
BLEU-4 & 0.1083 & 0.1110 & 0.329 & 0.1425 & $<$0.001\\
ROUGE-L & 0.285 & 0.274 & $<$0.001 & 0.301 & $<$0.001\\
CIDEr & 0.137 & 0.047 & $<$0.001 & 0.113 & 0.013\\
Stated findings, micro-F1 & 0.393 & 0.500 & $<$0.001 & 0.439 & $<$0.001\\
Stated findings, macro-F1 & 0.282 & 0.368 & 0.002 & 0.218 & $<$0.001\\
Clinical record, macro-F1 & 0.306 & 0.335 & 0.016 & 0.328 & 0.434\\
Clinical record, contradiction rate$\downarrow$ & 0.188 & 0.176 & 0.172 & 0.232 & $<$0.001\\
CRQS & 0.238 & 0.289 & 0.011 & 0.269 & 0.274\\
Hallucination rate$\downarrow$ & 0.453 & 0.469 & 0.187 & 0.439 & 0.038\\
Tokens per report & 139 & 167 & $<$0.001 & 143 & $<$0.001\\
\bottomrule
\end{tabular}

\end{table}

\noindent\textbf{Per variable (Table~\ref{tab:contentvar}).} For each variable, the reader's answers give whether a report
states it and, if so, whether the stated value agrees with the TCGA record (accuracy when stated). The decisions improve
what the slide shows: applied by rule to the retrieved report, they raise its histology macro-F1 from 0.35 to 0.49
($p=0.001$; paired bootstrap, 2{,}000 resamples) and its accuracy when stated from 0.73 to 0.81, and they do not make it
state the variables that H\&E does not determine more often (0.48 vs.\ 0.49 of reports). \method's histology (macro-F1
0.52; stated in 88\% of reports, 85\% accurate when stated) is above that of the retrieved report ($+0.17$, $p<0.001$),
of every slide-level LLM (0.18 to 0.29) and of every retrieval control (0.35 to 0.44), and below that of MI-Gen (0.55)
and \methodbase (0.58). \method's gain in clinical macro-F1 over \methodbase, in contrast, lies in the variables that the
diagnostic H\&E slide does not determine: it states them more often (0.48 $\rightarrow$ 0.65 of reports, $p<0.001$) at
the same accuracy (0.62 $\rightarrow$ 0.60, n.s.), so its macro-F1 on them rises (0.26 $\rightarrow$ 0.33) together
with its contradictions (Sec.~\ref{sec:clinical}), while it states tumour size, which the slide shows in part, less
often (23 vs.\ 43\% of reports; macro-F1 0.12 vs.\ 0.18). This gain is one of coverage, not of accuracy,
and on what H\&E cannot show more coverage is not desirable: with receptors withheld (Sec.~\ref{sec:abstain}), \method
states these variables less often than the references (0.36 vs.\ 0.48) at a similar accuracy (0.62).

\begin{table*}[!htb]
\caption{\textbf{Clinical variables by what the H\&E slide shows} (test; mean over seeds). Stated: share of reports that
state the variable; Acc.: share of stated values that agree with the TCGA record; F1: macro-F1 as in
Table~\ref{tab:main} (``not stated'' counted as a miss); each averaged over the variables of the group. Contr.:
contradiction rate over all seven variables.}
\label{tab:contentvar}
\centering\footnotesize
\setlength{\tabcolsep}{4pt}
\begin{tabular}{@{}lcccccccccc@{}}
\toprule
 & \multicolumn{6}{c}{Seen on the H\&E slide} & \multicolumn{3}{c}{Not determined by the H\&E slide} & \\
\cmidrule(lr){2-7}\cmidrule(lr){8-10}
 & \multicolumn{3}{c}{Histology} & \multicolumn{3}{c}{Tumour size} & \multicolumn{3}{c}{ER, PR, HER2, nodes, margin} & \\
\cmidrule(lr){2-4}\cmidrule(lr){5-7}\cmidrule(lr){8-10}
Reports & Stated & Acc. & F1 & Stated & Acc. & F1 & Stated & Acc. & F1 & Contr.$\downarrow$\\
\midrule
Reference reports & 0.74 & 0.90 & 0.55 & 0.58 & 0.84 & 0.59 & 0.48 & 0.91 & 0.49 & 0.052\\
\midrule
\methodbase & 0.91 & 0.88 & 0.58 & 0.43 & 0.53 & 0.18 & 0.48 & 0.62 & 0.26 & 0.169\\
\quad + length control & 0.89 & 0.89 & 0.57 & 0.44 & 0.53 & 0.18 & 0.56 & 0.63 & 0.31 & 0.189\\
\textbf{\method (ours)} & 0.88 & 0.85 & 0.52 & 0.23 & 0.45 & 0.12 & 0.65 & 0.60 & 0.33 & 0.232\\
\quad receptors withheld & 0.90 & 0.86 & 0.53 & 0.24 & 0.45 & 0.12 & 0.36 & 0.62 & 0.18 & 0.130\\
\midrule
Hospital consensus & 0.89 & 0.78 & 0.44 & 0.66 & 0.40 & 0.23 & 0.62 & 0.59 & 0.34 & 0.267\\
Consensus of 30 similar & 0.77 & 0.69 & 0.35 & 0.53 & 0.51 & 0.33 & 0.59 & 0.63 & 0.32 & 0.227\\
\midrule
Retrieved report (\psd), unedited & 0.73 & 0.73 & 0.35 & 0.60 & 0.55 & 0.35 & 0.48 & 0.66 & 0.29 & 0.188\\
\quad + \ccd/\csp edits by rule & 0.85 & 0.81 & 0.49 & 0.62 & 0.51 & 0.33 & 0.49 & 0.70 & 0.30 & 0.176\\
\midrule
TITAN + LLM (7B) & 0.54 & 0.75 & 0.18 & 0.88 & 0.58 & 0.26 & 0.38 & 0.70 & 0.16 & 0.152\\
\quad + retrieved reports as text & 0.81 & 0.83 & 0.45 & 0.80 & 0.56 & 0.30 & 0.50 & 0.66 & 0.28 & 0.193\\
\quad + \method decisions as text & 0.89 & 0.80 & 0.52 & 0.80 & 0.54 & 0.27 & 0.52 & 0.64 & 0.30 & 0.210\\
\bottomrule
\end{tabular}

\end{table*}

\noindent\textbf{Retrieved reports instead of decisions (Table~\ref{tab:llmdecisions}).} TITAN + LLM is trained exactly as
with \method's decisions (\supp{app:llm}; three seeds), but its prompt contains the cleaned reports of the five most
image-similar training cases instead of the decisions. The retrieved text gives the LLM much of what the decisions give:
its clinical macro-F1 rises from 0.175 to 0.310 (with the decisions: 0.324, difference n.s.), and its ROUGE-L, BERTScore
and REG score are higher than with the decisions. The decisions, however, determine which findings it states: with them,
its stated findings agree better with the reference (micro-F1 0.457 vs.\ 0.418, $p=0.019$; macro-F1 0.321 vs.\ 0.270,
$p=0.006$), so does its histology (macro-F1 0.52 vs.\ 0.45, $p=0.046$; Table~\ref{tab:contentvar}), and BLEU-1 and
METEOR are higher. \method, with 35M parameters, is ahead of the LLM given the retrieved reports in BLEU-1 to BLEU-4,
METEOR, ROUGE-L and the REG score (all $p\le0.03$) and in histology (0.52 vs.\ 0.45, $p=0.02$), at a similar clinical
macro-F1 (0.328 vs.\ 0.310, $p=0.11$) and with more contradictions (0.232 vs.\ 0.193, $p=0.005$), which explicit
abstention lowers to 0.130 (Sec.~\ref{sec:clinical}).

\section{Calibrated Claim Decision: decision and realisation}
\label{app:claims}

All numbers in this section are on the validation split (85 patients, seed 9233, validation-selected checkpoints); the
extractor is the one of the clinical-efficacy metric.

\noindent\textbf{Decision.} The global threshold chosen on validation is $\tau=0.17$ (0.19 and 0.27 for the slide-level models
of seeds 1 and 2). The decided claims have micro-F1 0.518 against the reference findings (test: 0.499) and number 5.7
per report, against 3.9 findings in the references: the F1-optimal rule accepts false positives to recover rare findings.
On the training split the mean calibrated posterior equals the prevalence (0.171 vs.\ 0.172), as expected from the
analytic correction of the class weighting; on validation it is lower than the prevalence (0.158 vs.\ 0.193), because
validation reports state more findings than training reports (3.9 vs.\ 3.4 per report).

\noindent\textbf{Realisation.} Table~\ref{tab:realise} measures how faithfully each decoder writes the claims it is given.
The claim attention raises the share of claimed findings that are written from 49\% (\methodbase + \psd, which does not read claims) to
58\% and lowers the share of unclaimed findings that are written from 9.7\% to 3.9\%. Given the reference's own findings as
claims, however, the same decoder writes only 57\% of them.

\begin{table}[!htb]
\caption{Claim realisation (validation). ``Claimed'' / ``unclaimed'': share of claimed / unclaimed (patient, finding)
pairs that the generated report states. Agreement: micro-F1 between written findings and claims. CE-F1: against the
reference. Oracle: the reference findings are given as claims (placed in the prior plan for \methodce).}
\label{tab:realise}
\centering\footnotesize
\setlength{\tabcolsep}{4pt}
\begin{tabular}{@{}lccccc@{}}
\toprule
 & Claimed & Unclaimed & Agreement & CE-F1 & CE-F1, oracle claims\\
\midrule
\methodbase + \psd & 0.491 & 0.097 & 0.566 & 0.422 & n/a\\
\methodbase + \psd + \ccd & 0.579 & 0.039 & 0.691 & 0.446 & 0.621\\
\methodce (+ \csp) & 0.530 & 0.031 & 0.659 & 0.477 & 0.590\\
\bottomrule
\end{tabular}
\end{table}

\section{Slide Knowledge Encoder and Plan-Conditioned Decoder: implementation and cost}
\label{app:backbone}

\begin{table}[!htb]
\caption{Notation.}
\label{tab:notation}
\centering\footnotesize
\setlength{\tabcolsep}{3pt}
\begin{tabular}{@{}ll@{}}
\toprule
Symbol & Meaning\\
\midrule
$\mathbf{X}\in\mathbb{R}^{N\times d_v}$ & tile embeddings of a slide (UNI2-h, $d_v{=}1536$)\\
$\mathbf{I},\mathbf{I}'\in\mathbb{R}^{N\times d}$ & projected tiles; output of the finding-aware tile attention\\
$\mathcal{T}$, $N_T$, $\mathbf{T}$ & topic bag (most frequent findings), $N_T{=}20$; its word embeddings\\
$\hat{\mathbf{T}},\hat{\mathbf{I}}\in\mathbb{R}^{N\times d}$ & findings attended by every tile; tiles gathered by them\\
$\mathbf{G}_{\mathrm{Pr}},\mathbf{W}_{\mathrm{Pr}}$ & finding-graph nodes; embeddings of the $N_K{=}100$ retrieved reports\\
$\mathbf{A},\hat{\mathbf{A}}$, $\mathbf{F}^{(l)}$ & adjacency of the finding graph, its normalisation; node features after $l$ graph convolutions\\
$\mathbf{G}',\mathbf{W}'\in\mathbb{R}^{N\times d}$ & knowledge-graph and retrieved-report encodings\\
$\mathbf{h}_t$ & decoder state at step $t$\\
$C$ & number of sentence types learned from training reports, $C{=}24$\\
$\bm{\pi}=(\pi_1,\dots,\pi_S)$, $L$ & report plan (types of its $S$ sentences), report length in tokens\\
$(\hat{\bm{\pi}},\hat L)$ & Exemplar Structure Prior from retrieved training reports (\psd)\\
$\sigma_t,r_t,c_t,\ell_t$ & sentences closed, remaining sentences, current type, remaining tokens\\
$\mathcal{N}_K(\mathbf{X})$ & $K$ most image-similar training cases\\
$z_j,\hat z_j$ & finding $j$ stated in the reference / claimed (\ccd)\\
$\tilde p_j$, $\tau$ & calibrated slide-level posterior; claim threshold\\
$\mathbf{C}\in\mathbb{R}^{N_T\times d}$ & claim tokens read by the decoder\\
$\hat{\mathbf{a}}_i$ & findings placed in planned sentence $i$ (\csp); ties go to the earliest sentence\\
$\mathbf{V}\in\mathbb{R}^{d\times N_T}$ & learned embedding of $\hat{\mathbf{a}}_{\sigma_t+1}$ in Eq.~\eqref{eq:esp}; $\hat{\mathbf{a}}_{\sigma_t+1}{=}\mathbf{0}$ after the plan\\
\bottomrule
\end{tabular}
\end{table}

\noindent\textbf{Inputs of the \ske.} The topic embedding $\mathbf{t}_j$ is the mean word embedding of the name of
finding $j$, taken from the decoder's embedding table and scaled by $\sqrt{d}$ like the decoder inputs. Retrieval ranks
the training cases by the cosine similarity of mean UNI2-h tile embeddings and never returns the query; the same ranking
gives the exemplars of the \psd (Sec.~\ref{sec:esp}). Each retrieved report is embedded once by a frozen BERT with
max-pooling over its outputs~\cite{devlin2019bert,liu2021exploring} and projected to $d$. The finding graph is described
under \emph{Finding graph} below.

\noindent\textbf{Per-tile fusion weights.} The weights of Eq.~\eqref{eq:distil} are
$[\bm{\lambda}_{1,t},\bm{\lambda}_{2,t}]=\sigma\big(\mathbf{h}_t\mathbf{W}_h\oplus\mathbf{\Gamma}\big)$ with
$\mathbf{\Gamma}=\mathbf{I}'\mathbf{W}_I+\mathbf{G}'\mathbf{W}_G+\mathbf{W}'\mathbf{W}_W\in\mathbb{R}^{N\times2}$, where
$\oplus$ adds the step term to every tile's row, and $\bm{\lambda}_{\cdot,t}$ is broadcast over the $d$ channels. Because the
key and value projections of the attention are linear, $\mathrm{MHA}(\mathbf{h}_t,\mathbf{I}'+\bm\lambda_1\odot\mathbf{G}'
+\bm\lambda_2\odot\mathbf{W}')$ is computed from the three projected sources and the per-tile weights without forming the
step-wise memory.

\noindent\textbf{Complexity.} Relative to \methodbase, \method adds $(R{+}C{+}4)d+N_Td$ parameters for the structure and
claim inputs and one claim attention per decoder layer ($4d^2{+}6d$ each), and per decoding step $O(N_Td)$ work plus a
counter and three look-ups. Per slide, the medoid plan costs $O(K^2S^2)$, the placement $O(N_TS)$ and the claims one
forward pass of the slide-level model; \sast adds no parameters. The cost of \methodbase is dominated by the tile-to-tile
attention of the \ske, $O(N^2d)$ time and, with fused attention kernels, $O(Nd)$ memory per layer,
and by the decoder's cross-attention, $O(TNd)$ per report (Table~\ref{tab:cost}).

\begin{algorithm}[t]
\caption{\method: report generation for one slide (inference)}
\label{alg:clasp}
\begin{algorithmic}[1]
\Require tile embeddings $\mathbf{X}$; training reports $\{y^{(j)}\}$, plans $\{\bm{\pi}^{(j)}\}$, statistics $P(j\mid k)$;
slide-level model $f_\phi$, threshold $\tau$; decoder $\theta$; beam $B$
\Ensure report $\hat y$
\State $\mathbf{I}'\gets\mathrm{TileAttn}(\mathbf{X}\mathbf{W}_x,\mathbf{T})$;\; $\mathcal{N}_K\gets$ $K$ most image-similar training cases \Comment{Eq.~\eqref{eq:tenc}}
\State $\mathbf{W}',\mathbf{G}'\gets\mathrm{KnowAttn}(\mathbf{I}',\mathbf{W}_{\mathrm{Pr}},\mathbf{G}_{\mathrm{Pr}})$ \Comment{Eq.~\eqref{eq:kenc}}
\State $(\hat{\bm{\pi}},\hat L)\gets$ medoid plan of $\mathcal{N}_K$, mean length of $\mathcal{N}_{K'}$ \Comment{Eq.~\eqref{eq:prior}}
\State $\hat z\gets\mathbb{1}[\sigma(f_\phi(\mathbf{X})-\log\mathbf{w})\ge\tau]$ \Comment{Eq.~\eqref{eq:claims}}
\State $\hat{\mathbf{a}}_{1:S}\gets$ place every claim in its best-fitting planned sentence \Comment{Eq.~\eqref{eq:cplan}}
\State $\mathbf{C}\gets$ claim tokens of $\hat z$;\; beams $\gets\{(\langle\mathrm{bos}\rangle,0)\}$
\For{$t=0,1,\dots$ \textbf{until} every beam has emitted $\langle\mathrm{eos}\rangle$}
  \ForAll{beam $(y_{\le t},s)$}
    \State $\sigma_t\gets$ sentences closed in $y_{\le t}$;\; $r_t,c_t,\ell_t,\hat{\mathbf{a}}_{\sigma_t+1}$ from the plan
    \State $\mathbf{e}_t\gets$ word + position + structure embedding \Comment{Eq.~\eqref{eq:esp}}
    \State every layer: adaptive knowledge fusion of $\mathbf{I}',\mathbf{G}',\mathbf{W}'$, then claim attention over $\mathbf{C}$ \Comment{Eqs.~\eqref{eq:distil},\eqref{eq:claimattn}}
    \State scores of all continuations $s+\log p_\theta(\cdot\mid\cdot)$
  \EndFor
  \State keep the $B$ best continuations (trigram blocking, no immediate repeats)
\EndFor
\State \Return the best finished beam
\end{algorithmic}
\end{algorithm}

\noindent Lines 1 and 2 are the \ske. Lines 3 to 6 are the three decisions of \method, made once per slide
before decoding: the structure from the reports of similar cases (line 3), the claims from the calibrated slide-level
posteriors (line 4) and the placement of every claim in a planned sentence (line 5). Lines 7 to 16 are standard beam search;
the decisions enter in lines 9 to 11 through a counter of written sentences, four embedding look-ups and one attention over
$N_T$ claim tokens per layer. The total cost is that of \methodbase, $O(N^2d)$ for the finding-aware tile attention and $O(TNd)$ for the decoder's
cross-attention, plus $O(TN_Td)$ for the claim attention and one slide-level pass per slide.

\noindent\textbf{Size and cost (Table~\ref{tab:cost}).} \method has 35.1M parameters, including the frozen
2.4M-parameter slide-level model, and writes a report in 1.6\,s with 0.51\,GiB of peak GPU memory. The three decisions add
5.6M parameters and 3\% of the compute to \methodbase; \sast adds nothing at inference. MI-Gen and HistGen are smaller
(15.2M, 22.0M); RANGER reports neither its size nor its run time.

\begin{table}[t]
\caption{\textbf{Size and cost.} Parameters of the forward pass on tile embeddings (\method: with the frozen slide-level model; \sast adds none) and fp32 size; GFLOPs of one teacher-forced pass ($10^4$ tiles, 150 tokens); latency and peak memory of full inference per slide (bf16, RTX A5000, mean of 20 test slides). n.r.: not reported.}
\label{tab:cost}
\centering\footnotesize
\setlength{\tabcolsep}{2.2pt}
\begin{tabular}{@{}lccccc@{}}
\toprule
Model & Params & Size & GFLOPs & s/slide & Peak GiB\\
\midrule
MI-Gen~\cite{chen2024wsicaption} & 15.2M & 61\,MB & n/a & n/a & n/a\\
HistGen~\cite{guo2024histgen} & 22.0M & 88\,MB & n/a & n/a & n/a\\
RANGER~\cite{chen2026ranger} & n.r. & n.r. & n.r. & n.r. & n.r.\\
\midrule
\methodbase & 29.5M & 118\,MB & 570 & 1.28 & 0.50\\
\textbf{\method (ours)} & 35.1M & 140\,MB & 587 & 1.60 & 0.51\\
\bottomrule
\end{tabular}
\end{table}

\noindent\textbf{Sources of every setting.} The architecture uses $d{=}512$, 8 heads, FFN width $4d$, $N_T{=}20$ topics,
$N_K{=}100$ retrieved reports, BERT with max-pooling for report embeddings, multi-label pre-training with a weighted binary
cross-entropy, and Adam with $\beta{=}(0.8,0.999)$, learning rate $10^{-4}$ and batch size 16. The Transformer settings
follow R2Gen~\cite{chen2020r2gen} (3 decoder layers, dropout 0.1, Xavier initialisation, gradient clipping at 0.1), the
graph module follows~\cite{zhang2020kg} (graph design, node initialisation, residual graph convolutions, auxiliary
per-node loss) with our breast compartments in place of their chest groups, and the
data and decoding settings follow the PathText benchmark code~\cite{chen2024wsicaption} (vocabulary threshold 3, maximum report length 600, beam 3 with trigram
blocking, validation every 2 epochs from epoch 10, early stopping after 20 epochs without improvement of validation BLEU-4).

\noindent\textbf{Topic bag.} The 20 topics are the most frequent findings of the training reports under a
rule-based extractor with negation handling (synonym patterns for 46 breast-pathology findings); ``high nuclear grade''
is skipped as redundant with histologic grade 3.

\begin{table}[!htb]
\caption{Compartments of the finding graph (Sec.~\ref{sec:backbone}). The findings of a compartment are connected pairwise;
the global node is connected to all 20 findings. In parentheses: share of training reports that state the finding.}
\label{tab:graph}
\centering\footnotesize
\begin{tabular}{@{}lp{0.62\linewidth}cc@{}}
\toprule
Compartment & Findings & Pairs & Degree\\
\midrule
Invasive tumour & invasive ductal carcinoma (59.0\%), histologic grade 2 (26.3\%), histologic grade 3 (26.0\%),
lymphovascular invasion (20.6\%), estrogen receptor positive (15.4\%), invasive lobular carcinoma (13.8\%), tumour
necrosis (11.9\%), multifocal tumour (9.6\%) & 28 & 8\\
Ducts and lobules & ductal carcinoma in situ (31.9\%), microcalcifications (15.9\%), lobular carcinoma in situ (9.1\%),
usual ductal hyperplasia (8.8\%), atypical ductal hyperplasia (6.2\%) & 10 & 5\\
Benign parenchyma & fibrocystic change (20.1\%), microcalcifications (15.9\%), biopsy site changes (12.9\%), stromal
fibrosis (6.2\%), sclerosing adenosis (5.6\%) & 10 & 5\\
Axillary lymph nodes & lymph node metastasis (29.4\%), extranodal extension (9.8\%) & 1 & 2\\
Surgical margins & positive surgical margin (5.8\%) & 0 & 1\\
\midrule
\multicolumn{2}{@{}l}{Global node, connected to every finding} & 20 & 20\\
\bottomrule
\end{tabular}
\end{table}

\noindent\textbf{Finding graph.} The graph follows Zhang~\etal~\cite{zhang2020kg}: the findings are its nodes, findings
that concern the same part of the specimen are connected by bidirectional edges, and a global node is connected to all of
them. Zhang~\etal group chest findings by organ or body part; the five breast compartments of Table~\ref{tab:graph},
and the assignment of each finding to them, are ours. Microcalcifications belong to two compartments (degree 9). Pairs within compartments (49) and the edges of the
global node (20) give 69 undirected edges, that is, 138 non-zero entries of the symmetric matrix $\mathbf{A}$, which has no
self-loops. With $\hat{\mathbf{A}}=\mathbf{D}^{-1/2}\mathbf{A}\mathbf{D}^{-1/2}$, an edge between nodes of degrees
$d_u$ and $d_v$ has weight $1/\sqrt{d_ud_v}$: 0.125 between two invasive findings, 0.5 between lymph node metastasis and
extranodal extension, and 0.079 between an invasive finding and the global node, so the global node, linked to all findings, contributes less to a finding than any neighbour of its own compartment. The positive surgical margin has no
compartment neighbour and is linked to the global node only. Through the global node every finding is within two edges of
every other, so two graph convolutions let information reach all nodes; the compartments decide which findings exchange
information directly and with the largest weights. The initial node features are pooled from the
projected tiles,
\begin{equation}
\mathbf{F}^{(0)}=\big[\operatorname{softmax}_{N}(\mathbf{I}\mathbf{W}_a)^{\top}\mathbf{I};\;\tfrac{1}{N}\mathbf{1}^{\top}\mathbf{I}\big]\in\mathbb{R}^{(N_T+1)\times d},
\label{eq:gnode}
\end{equation}
where the softmax normalises each column of $\mathbf{I}\mathbf{W}_a\in\mathbb{R}^{N\times N_T}$ over the $N$ tissue tiles,
so that finding $j$ (column $\mathbf{w}_j$ of $\mathbf{W}_a$) is pooled from the tiles that resemble it, and the last row
is the global node. Layer $l\in\{0,1\}$ of the graph convolution computes
\begin{equation}
\mathbf{M}^{(l)}=\mathrm{ReLU}\big(\mathrm{BN}(\hat{\mathbf{A}}\mathbf{F}^{(l)}\mathbf{W}_m^{(l)})\big),\qquad
\mathbf{F}^{(l+1)}=\mathbf{F}^{(l)}+\mathrm{ReLU}\big(\mathrm{BN}([\mathbf{F}^{(l)},\mathbf{M}^{(l)}]\,\mathbf{W}_u^{(l)})\big),
\label{eq:gcn}
\end{equation}
where $\mathbf{M}^{(l)}$ is the message that each node collects from its neighbours, $[\cdot,\cdot]$ concatenates the
channels, $\mathbf{W}_m^{(l)}\in\mathbb{R}^{d\times d}$ and $\mathbf{W}_u^{(l)}\in\mathbb{R}^{2d\times d}$ are linear
maps with bias, and the residual keeps each node's own tile evidence. $\mathbf{G}_{\mathrm{Pr}}$ is formed by the first
$N_T$ rows of $\mathbf{F}^{(2)}$.

\noindent\textbf{Training of the graph.} Stage 1 trains the tile projection and the graph module, as in
Zhang~\etal~\cite{zhang2020kg}, with a weighted binary cross-entropy on the 20 silver topic labels, applied to a linear
classifier on the mean of the 21 final node features and to an auxiliary per-node classifier on the initial finding
features $\mathbf{F}^{(0)}$ (early stopping on validation macro-AUC; 0.713/0.695/0.705 for the three seeds). Stage 2
trains the whole model, including the graph, with the token-level cross-entropy of the reference report, conditioned on
the target's own structure and findings (Secs.~\ref{sec:esp} to~\ref{sec:cplan}); the running statistics of the graph's
batch normalisation are frozen at their stage-1 values, while its scale and shift are trained. Zhang~\etal instead keep
the classifier fixed during report generation, and describe the propagation with a normalised graph Laplacian; we use the
normalised adjacency above.

\noindent\textbf{Exact adaptive knowledge fusion.} Adaptive knowledge fusion mixes the memory per decoding step and per tile,
$\mathbf{M}_t[j]=\mathbf{I}'_j+\lambda_{1,t,j}\mathbf{G}'_j+\lambda_{2,t,j}\mathbf{W}'_j$, which cannot be materialised for
$10^4$ tiles and hundreds of steps. Because keys and values are linear in the memory, the attention scores and outputs
decompose exactly into three standard attention products,
$\mathbf{q}_t\mathbf{K}_t^\top=\mathbf{q}_t\mathbf{K}_{I}^\top+\bm{\lambda}_{1,t}\odot\mathbf{q}_t\mathbf{K}_{G}^\top+\bm{\lambda}_{2,t}\odot\mathbf{q}_t\mathbf{K}_{W}^\top$
(bias terms folded into $\mathbf{K}_I$), verified equal to the materialised form to $2\cdot10^{-16}$.

\noindent\textbf{Bags and batches.} A slide contributes all its tissue tiles (UNI2-h, 256\,px at 20$\times$; mean 12{,}016,
median 11{,}573, 95th percentile 22{,}497). Bags larger than 10{,}000 tiles (the benchmark's cap; 61\% of slides) are
subsampled uniformly at random in every training epoch and at evenly spaced indices at evaluation; because tiles are stored in
raster order, the latter is spatially uniform (the benchmark code keeps the first 10{,}000 rows, which would drop the lower part
of large slides). A mini-batch of 16 patients is packed into as few forward passes as fit a memory budget; padded tiles are
masked and the loss is normalised by the number of target tokens of the whole mini-batch, which gives exactly the gradient of
an unpacked batch (verified in float64, relative error $10^{-15}$).

\noindent\textbf{What adaptive knowledge fusion does.} Averaged over the tokens of generated test reports, the fusion weights of \methodbase
(seed 9233) are $\lambda_1\le1.9{\cdot}10^{-4}$ (graph) and $\lambda_2\le3.1{\cdot}10^{-8}$ (retrieved reports) for both
abnormality and normality sentences: the model effectively switches the graph and the retrieved reports off, and
\methodbase reduces to the finding-aware tile attention and the decoder. Retrieved reports therefore reach \method only through
the \psd (Sec.~\ref{sec:esp}).

\section{Generalisation to PathText-LUAD}
\label{app:luad}

\noindent\textbf{Data.} Reports: the PathText reports of TCGA-LUAD. Tile embeddings: UNI2-h embeddings from the same
public feature release as for BRCA, of the first diagnostic slide of every patient with a report (466 patients;
12{,}035 tiles per slide on average, 55\% of the slides above the $10^4$-tile cap). Split: random at the patient level
(seed 0) with the proportions of BRCA, 380/42/44 patients. The 33 contributing sites do not overlap with the 38 sites of
PathText-BRCA. All report-derived quantities (vocabulary of 855 words, sentence types, retrieval corpus, finding topics,
claim-placement statistics) use training reports only.

\noindent\textbf{Findings.} The concepts of the breast extractor do not apply to the lung, so we wrote lung concepts with
the same matching and negation rules; the breast concepts and every BRCA output are unchanged. The 20 most frequent
findings of the training reports form the topic bag: adenocarcinoma; acinar, papillary, micropapillary, lepidic and
mucinous patterns and a squamous component; well, moderately and poorly differentiated; tumour necrosis; multifocal
tumour (graph group \emph{tumour}); visceral pleural and lymphovascular invasion (\emph{invasion}); lymph-node metastasis
(\emph{lymph nodes}); granuloma, emphysema, fibrosis, inflammation and anthracosis (\emph{background lung}).

\noindent\textbf{Settings.} Model, training and decoding settings are those of BRCA (Sec.~\ref{app:config}), and
the protocol was fixed before training. Each of the three seeds has its own slide-level model; the claim threshold is
selected on LUAD validation by the rule used for BRCA ($\tau=0.37, 0.52, 0.27$; test claim micro-F1 0.45, 0.43, 0.55).
Checkpoints are selected on validation BLEU-4: \sast stops at epochs 3, 6 and 7 of at most 8, with validation BLEU-4
0.146, 0.143 and 0.148 against 0.134, 0.131 and 0.131 for \methodce. The test split was decoded once per model, after all
training had ended.

\begin{table}[!htb]
\caption{\textbf{Generalisation to PathText-LUAD} (33 sites, none shared with BRCA; 44 test patients). Mean
\pmstd{std} over three seeds; F1$_\mathrm{f}$: F1 of the stated findings; $^{*}$/$^{\dagger}$: \method better/worse,
$p<0.05$.}
\label{tab:luad}
\centering\footnotesize
\setlength{\tabcolsep}{5pt}
\begin{tabular}{@{}lcccccc@{}}
\toprule
Method & B-1 & B-4 & MTR & R-L & CIDEr & F1$_\mathrm{f}$\\
\midrule
Most similar training report & \second{0.430} & 0.095$^{*}$ & \best{0.196}$^{\dagger}$ & 0.284 & \best{0.143} & 0.449\\
Consensus of 30 similar & 0.412$^{*}$ & 0.091$^{*}$ & \second{0.170}$^{\dagger}$ & 0.270$^{*}$ & 0.088 & 0.458\\
\midrule
\methodbase & 0.363\pmstd{0.021}$^{*}$ & 0.097\pmstd{0.007}$^{*}$ & 0.141\pmstd{0.006}$^{*}$ & \second{0.286}\pmstd{0.002}$^{*}$ & 0.086\pmstd{0.031} & \best{0.479}\pmstd{0.030}\\
\methodce & 0.418\pmstd{0.002}$^{*}$ & \second{0.115}\pmstd{0.002}$^{*}$ & 0.149\pmstd{0.002} & \second{0.286}\pmstd{0.007}$^{*}$ & \second{0.123}\pmstd{0.010} & 0.417\pmstd{0.046}$^{*}$\\
\textbf{\method (ours)} & \best{0.434}\pmstd{0.007} & \best{0.128}\pmstd{0.004} & 0.152\pmstd{0.001} & \best{0.297}\pmstd{0.007} & 0.116\pmstd{0.007} & \second{0.469}\pmstd{0.039}\\
\bottomrule
\end{tabular}
\end{table}

\noindent\textbf{Paired differences} (\method minus the other system, 95\% patient-bootstrap interval). Against
\methodce: BLEU-4 $+0.013$ $[0.007, 0.019]$, ROUGE-L $+0.011$ $[0.005, 0.018]$, F1 of the stated findings $+0.052$
$[0.012, 0.094]$. Against \methodbase: BLEU-4 $+0.031$ $[0.022, 0.038]$, METEOR $+0.010$ $[0.005, 0.016]$, ROUGE-L
$+0.011$ $[0.002, 0.020]$, F1 of the stated findings $-0.010$ $[-0.051, 0.031]$. CIDEr differs significantly from neither. BLEU-2 and BLEU-3, not shown in Table~\ref{tab:luad}, are 0.291 and 0.195 for \method, 0.277 and 0.181 for \methodce, 0.235 and 0.152 for \methodbase and 0.255 and 0.150 for the most similar training report.
The clinical-record evaluation of Sec.~\ref{sec:setup} reads breast-specific variables (receptor status, ductal and
lobular histology) and is not applied to this cohort.

\section{Re-training the baselines}
\label{app:baselines}

\noindent\textbf{Principle.} We keep the public code of each baseline unchanged and replace only the data interface, so
that every method sees the same split, the same tile embeddings with the same subsampling of large bags, the same report
text and the same vocabulary. Model, loss, optimiser, schedule, decoding and checkpoint selection on validation BLEU-4 are
those of the original code, and all generated reports are scored by the same evaluation code against identical references.

\noindent\textbf{MI-Gen.} We build the vocabulary from training reports only (the original code builds it from all
reports, including the test reports) and feed the split, reports and 1536-d UNI2-h tile embeddings of our protocol. Two
changes concern efficiency only: the attention, which materialises the $N{\times}N$ score matrix and does not fit a
24\,GB GPU at $10^4$ tiles, uses a fused attention kernel with the same mask and dropout, and the test split is not
decoded during training.

\noindent\textbf{HistGen.} The original tokenizer applies its own text cleaning, which removes sentence periods, so its
outputs could not be scored against the benchmark references on equal terms; we use the benchmark cleaner and the
training vocabulary instead. The input projection, designed for 1024-d features, is re-created for 1536-d UNI2-h
features with the same initialisation, and the per-epoch test decoding is skipped; training and validation are otherwise
unchanged. We use the hyper-parameters of HistGen's training script (40 epochs, batch size 1, learning rate $10^{-4}$,
step decay 0.8, region size 96, 512 prototypes, beam 3) except two data-protocol values shared by all methods: the
script truncates reports to 100 tokens, which would cut 73\% of the PathText-BRCA training reports, so we use the
benchmark's 600 tokens, and its vocabulary threshold of 10 is replaced by the common training vocabulary (threshold 3).
The truncation also affects evaluation in the original code, which scores generated reports against the first 100
tokens of each reference, whereas the released code of BiGen and MI-Gen uses 600 tokens. The HistGen numbers quoted in
Table~\ref{tab:main} come from BiGen's re-run, whose settings are not stated. Under our protocol, HistGen's
validation-selected model (epoch 11 of 40) writes reports of 94 tokens on average on the test split; our scorer reproduces
its own validation BLEU-4 exactly.

\noindent\textbf{BiGen.} Not re-trained: it concatenates UNI and PLIP embeddings of the same 10$\times$ patches and
retrieves PLIP sentence embeddings, so it requires PLIP features of the tiles we use. Two discrepancies between the paper
and the released code are relevant for reproduction: the region size $m$ is 20 in the paper and 10 in the code, and the
code selects checkpoints on validation ROUGE-L rather than BLEU-4.

\noindent\textbf{Seeds.} Both re-trained baselines use one seed, the default of their code. MI-Gen's decoder recomputes
all layers over the whole prefix at every step and needs about an hour per validation round on our shared GPUs, so three
seeds per baseline were not affordable; our models are reported over three seeds.

\noindent\textbf{Test-selected records in public code.} The training code of WsiCaption, BiGen and HistGen scores the
test split at every validation round and records, besides the validation-selected result, the best test score over
epochs. Which record a paper reports is not stated. In a re-run of \methodbase that scored the test split at every
validation round (scores logged and used for nothing else), test BLEU-4 ranged from 0.079 to 0.121 over the 18 rounds,
the size of the gaps between published methods; the validation-selected round happened to be the best test round.

\noindent\textbf{Hospital consensus.} A retrieval rule without learned parameters: the predicted contributing site is
the most frequent tissue source site among the $k_h$ most image-similar training slides (read from the training
patients' TCGA identifiers; the query's own identifier is never used), and the output is the report with the highest mean
sentence BLEU-4 to the others among the $n$ most image-similar training reports of that site. Over the grid
$k_h\in\{1,5,10,20\}$, $n\in\{10,30,\text{all}\}$, validation BLEU-4 is highest for $k_h{=}1$ and $n{=}$all (0.126; the
site is correct for 91.8\% of validation and 82.6\% of test patients), and the test split was decoded once with this
setting.

\noindent\textbf{SCOUT and PathReportEval.} SCOUT~\cite{singh2026scout} and the BiGen adaptation of
PathReportEval~\cite{singh2026pathreporteval} are trained and tested with the public code of PathReportEval (PathBench,
commit a110a98), whose TCGA split pools all TCGA projects (7{,}074/884/885 training/validation/test cases). Matched by
patient identifier, 75 of our 92 test patients are training cases of that split and 10 are validation cases (of our 85
validation patients, 73 and 7). Scores reported on that split are therefore not comparable with Table~\ref{tab:main},
and a model trained on it has seen most of our test reports. Re-training SCOUT under our protocol also needs its
slide-level and concept features; we leave it to future work.

\section{Slide-level LLM baselines}
\label{app:llm}

\noindent\textbf{Why released checkpoints are not evaluated.} The instruction data of SlideChat and the WSI-Bench training
data of WSI-LLaVA are derived from TCGA pathology reports, and they contain the reports of 75 and of all 92 PathText-BRCA
test patients (Table~\ref{tab:contamination}). A score of these checkpoints on the test split would measure, in part,
recall of reports seen in training. PRISM~\cite{shaikovski2024prism} needs Virchow tile features of the slides and
access-restricted weights, and the weights of PathChat~\cite{lu2024pathchat}, an assistant for regions of interest, are
not released; neither was evaluated. We therefore re-train the published architectures on the PathText-BRCA training split
only, under exactly the protocol of \method.

\begin{table}[!htb]
\caption{\textbf{Released slide-level LLMs have seen the PathText-BRCA test reports.} Patients of the PathText-BRCA split
whose TCGA identifier occurs in the training or evaluation files of SlideChat~\cite{chen2025slidechat} (public Hugging
Face dataset) and WSI-LLaVA~\cite{liang2025wsillava} (WSI-Bench files of its public repository); both are built from TCGA
pathology reports. Script: \texttt{contamination\_check.py} of the release.}
\label{tab:contamination}
\centering\footnotesize
\setlength{\tabcolsep}{5pt}
\begin{tabular}{@{}llcccc@{}}
\toprule
Model & File & Patients & Train (792) & Val (85) & Test (92)\\
\midrule
SlideChat & SlideInstruction, stage-1 captions & 3294 & 602 & 65 & \textbf{75}\\
 & SlideInstruction, stage-2 VQA & 3294 & 602 & 65 & \textbf{75}\\
 & SlideBench-Caption (its test set) & 1547 & 147 & 14 & 11\\
\midrule
WSI-LLaVA & WSI-Bench, train & 8383 & 790 & 85 & \textbf{92}\\
 & WSI-Bench, test & 727 & 161 & 15 & 16\\
\bottomrule
\end{tabular}
\end{table}

\noindent\textbf{Common protocol.} Every LLM baseline sees only the training reports (791 for the SlideChat architecture:
its released features miss one training patient), is selected on validation BLEU-4 and decodes the test split once, with
three seeds. Targets and references are those of \method (cleaned reports mapped through the training vocabulary), and
generated text passes through the benchmark cleaner. Decoding is greedy, with no repeated trigrams, the
\textlangle unk\textrangle{} token suppressed as for \method, and at most 600 new tokens. Scoring uses the code and
environment of every other result; re-scoring the released \method predictions in this environment reproduces all
benchmark metrics, all 644 reader answers and all REG, CRQS and BERTScore values of seed 9233 exactly.

\noindent\textbf{HistoGPT~\cite{tran2025histogpt}.} The released HistoGPT-1B model (822M parameters: BioGPT decoder,
Perceiver resampler, gated cross-attention) is fine-tuned end to end with its report prompt ``Final diagnosis:''. As for
every re-trained baseline, all methods read the same UNI2-h tiles (at most $10^4$ per slide), so the input layer is
re-created: a LayerNorm and a linear map from 1536 to its 768 input dimensions. AdamW, learning rate $2\cdot10^{-5}$
($10^{-4}$ for the adapter), 8 epochs, validation every 2 epochs, batch 8 by gradient accumulation, bfloat16 autocast.

\noindent\textbf{LLM decoder.} The three other baselines share a decoder: Qwen2.5-7B-Instruct with LoRA ($r{=}16$,
$\alpha{=}32$, dropout 0.05) on every linear projection; AdamW, batch 8 by gradient accumulation, 3 epochs with validation
after each, linear warm-up over 3\% of the steps and linear decay, bfloat16, gradient checkpointing, one RTX A5000. The
learning rate was chosen once, on validation, with TITAN + LLM and seed 9233 ($2\cdot10^{-4}$: 0.0876 vs.\
$5\cdot10^{-5}$: 0.0675 validation BLEU-4) and used for every other run. The visual input differs:
(i)~\emph{SlideChat architecture}~\cite{chen2025slidechat}: SlideChat's released CONCH patch features (3.2k patches per
slide on average, at most 4096, evenly spaced as in its loader), its LongNet slide encoder (2 layers, 512-d) and its
two-layer projector, one token per patch; trained from scratch on our split.
(ii)~\emph{TITAN + LLM}: the TITAN~\cite{ding2025titan} slide embedding released by its authors for TCGA (768-d, first
diagnostic slide), mapped linearly to 4 prefix tokens.
(iii)~\emph{TITAN + LLM + \method decisions}: (ii) with \method's decisions as text in the prompt, \eg ``Report plan (9
sentences): T21, T4, T4, T2, T10 [Invasive Ductal Carcinoma; Ductal Carcinoma In Situ; Histologic Grade 2; \dots], \dots{}
Length: about 149 tokens.'' During training these are the target's own plan, findings per sentence and length, as for
\methodce; at validation and test they are the test-time decisions of the seed (\psd, \ccd, \csp).
(iv)~\emph{LLM without visual input}: the same LLM, prompt and training with no visual tokens, a control for what the
visual input adds to the LLM's prior over reports.
(v)~\emph{TITAN + LLM + retrieved reports}: (ii) with the cleaned reports of the five most image-similar training
cases (the \psd's $K'$, in rank order; for a training patient never its own report) as text in the prompt instead of the
decisions, a check of what the decisions add to the retrieved text (\supp{app:content}).

\noindent\textbf{What the LLMs write (Table~\ref{tab:llmdiag}).} Greedy decoding of a fixed prompt gives the LLM
without visual input one report for all 92 test patients, and this single report reaches the BLEU-4 of the
slide-conditioned LLMs (Table~\ref{tab:main}). These write 17 to 74 distinct reports for the 92 patients and state the
most frequent findings. Fine-tuned on 792 reports, the LLMs thus learn a typical report rather than a mapping from the
slide, the failure that Sec.~\ref{sec:analysis} identifies, amplified. HistoGPT, which reads every tile through
gated cross-attention, writes almost one report per patient but under-generates the most.

\begin{table}[!htb]
\caption{\textbf{What the re-trained baselines write} (test split): mean tokens per report, number of distinct reports
among the 92 test reports, and macro-F1 of the stated findings (20 findings of the rule-based extractor).}
\label{tab:llmdiag}
\centering\footnotesize
\setlength{\tabcolsep}{5pt}
\begin{tabular}{@{}lccc@{}}
\toprule
Method & Tokens per report & Distinct reports (seeds 9233 / 1 / 2) & Macro-F1 of stated findings\\
\midrule
HistoGPT-1B & 98 & 83 / 88 / 87 & 0.183\\
SlideChat arch.\ (7B) & 110 & 36 / 17 / 27 & 0.098\\
TITAN + LLM (7B) & 110 & 28 / 74 / 49 & 0.133\\
\quad + \method decisions as text & 134 & 92 / 92 / 91 & 0.321\\
LLM without visual input (7B) & 111 & 1 / 1 / 1 & 0.086\\
\method (ours) & 143 & 92 / 92 / 92 & 0.218\\
\midrule
Reference reports & 149 & 92 & n/a\\
\bottomrule
\end{tabular}

\end{table}

\noindent\textbf{Effect of the decisions (Table~\ref{tab:llmdecisions}).} With \method's decisions in its prompt, TITAN +
LLM writes patient-specific reports (91 to 92 distinct reports) of the planned length, and every overlap metric, the F1
of the stated findings, the agreement with the clinical record, CRQS and the REG score rise significantly, while the
hallucination rate falls. As for \sast in \method (Sec.~\ref{sec:clinical}), the contradiction
rate rises: the decided claims and the planned sentence types make the LLM state receptor status in 33 to 36\% of its
reports instead of 3 to 4\%, and the progesterone- and HER2-receptor contradictions rise from 0.01 to 0.17 and 0.16. This
is the variable that explicit abstention withholds.

\begin{table}[!htb]
\caption{\textbf{\method's decisions given to a 7B LLM.} TITAN + LLM without and with \method's decisions as text in
its prompt (means over three seeds; difference and one-sided $p$ of the paired bootstrap over test patients, 10{,}000
resamples, in the direction of the observed difference); right: the same LLM given the reports of the five most similar
training cases as text instead of the decisions (\supp{app:content}), and the LLM with \method's decisions minus it.}
\label{tab:llmdecisions}
\centering\footnotesize
\setlength{\tabcolsep}{3pt}
\begin{tabular}{@{}lccccccc@{}}
\toprule
Metric & TITAN + LLM & + \method decisions & Difference & $p$ & + retrieved reports & Decisions $-$ retrieved & $p$\\
\midrule
BLEU-1 & 0.340 & 0.405 & $+0.065$ & $<$0.001 & 0.351 & $+0.054$ & $<$0.001\\
BLEU-4 & 0.0780 & 0.0878 & $+0.0099$ & 0.004 & 0.0926 & $-0.0047$ & 0.148\\
METEOR & 0.140 & 0.161 & $+0.021$ & $<$0.001 & 0.153 & $+0.008$ & 0.033\\
ROUGE-L & 0.251 & 0.260 & $+0.009$ & 0.003 & 0.279 & $-0.019$ & $<$0.001\\
CIDEr & 0.039 & 0.076 & $+0.037$ & 0.003 & 0.092 & $-0.017$ & 0.267\\
Stated findings, micro-F1 & 0.342 & 0.457 & $+0.114$ & $<$0.001 & 0.418 & $+0.039$ & 0.019\\
Stated findings, macro-F1 & 0.133 & 0.321 & $+0.188$ & $<$0.001 & 0.270 & $+0.051$ & 0.006\\
Clinical record, macro-F1 & 0.175 & 0.324 & $+0.149$ & $<$0.001 & 0.310 & $+0.014$ & 0.217\\
Clinical record, contradiction rate$\downarrow$ & 0.152 & 0.210 & $+0.058$ & $<$0.001 & 0.193 & $+0.017$ & 0.135\\
CRQS & 0.207 & 0.248 & $+0.041$ & 0.047 & 0.241 & $+0.007$ & 0.391\\
Hallucination rate$\downarrow$ & 0.517 & 0.484 & $-0.033$ & 0.007 & 0.458 & $+0.026$ & 0.058\\
Clinical discordance$\downarrow$ & 0.331 & 0.314 & $-0.016$ & 0.101 & 0.302 & $+0.012$ & 0.266\\
REG score & 0.324 & 0.332 & $+0.009$ & $<$0.001 & 0.347 & $-0.015$ & $<$0.001\\
Tokens per report & 110 & 134 & $+24$ & $<$0.001 & 107 & $+27$ & $<$0.001\\
\bottomrule
\end{tabular}

\end{table}

\noindent\textbf{Checkpoint selection.} Table~\ref{tab:llmselection} lists the validation BLEU-4 of every validated epoch
of every baseline run, the selected epoch and the mean length of its test reports. The release contains the logs, the
validation predictions of the selected epoch, the test predictions and the weights (LoRA adapters and visual modules)
of every run.

\begin{table}[!htb]
\caption{\textbf{Checkpoint selection of the re-trained baselines.} Validation BLEU-4 (epoch in parentheses; bold:
selected) and mean number of tokens of the test reports (references: 149). The last row is the learning-rate trial
that was not chosen.}
\label{tab:llmselection}
\centering\footnotesize
\setlength{\tabcolsep}{4pt}
\begin{tabular}{@{}llcccccc@{}}
\toprule
Baseline & Seed & \multicolumn{4}{c}{Validation BLEU-4 per validated epoch} & Selected & Test length\\
\midrule
HistoGPT-1B & 9233 & 0.0649 (2) & 0.0674 (4) & \textbf{0.0804} (6) & 0.0721 (8) & 6 & 106\\
 & 1 & 0.0693 (2) & \textbf{0.0787} (4) & 0.0740 (6) & 0.0699 (8) & 4 & 93\\
 & 2 & 0.0734 (2) & \textbf{0.0830} (4) & 0.0760 (6) & 0.0818 (8) & 4 & 96\\
\midrule
SlideChat arch. & 9233 & \textbf{0.0973} (1) & 0.0490 (2) & 0.0519 (3) &  & 1 & 130\\
 & 1 & 0.0531 (1) & \textbf{0.0632} (2) & 0.0585 (3) &  & 2 & 86\\
 & 2 & \textbf{0.0735} (1) & 0.0564 (2) & 0.0689 (3) &  & 1 & 115\\
\midrule
TITAN + LLM & 9233 & \textbf{0.0876} (1) & 0.0423 (2) & 0.0632 (3) &  & 1 & 115\\
 & 1 & 0.0499 (1) & 0.0691 (2) & \textbf{0.0755} (3) &  & 3 & 108\\
 & 2 & 0.0407 (1) & \textbf{0.0716} (2) & 0.0635 (3) &  & 2 & 106\\
\midrule
TITAN + LLM + \method decisions & 9233 & 0.0864 (1) & \textbf{0.0954} (2) & 0.0854 (3) &  & 2 & 133\\
 & 1 & 0.0850 (1) & \textbf{0.0945} (2) & 0.0923 (3) &  & 2 & 135\\
 & 2 & 0.0865 (1) & 0.0806 (2) & \textbf{0.0887} (3) &  & 3 & 134\\
\midrule
LLM without visual input & 9233 & \textbf{0.0966} (1) & 0.0475 (2) & 0.0604 (3) &  & 1 & 139\\
 & 1 & 0.0499 (1) & \textbf{0.0697} (2) & 0.0664 (3) &  & 2 & 91\\
 & 2 & 0.0627 (1) & 0.0682 (2) & \textbf{0.0708} (3) &  & 3 & 102\\
\midrule
TITAN + LLM, lr $5\cdot10^{-5}$ & 9233 & 0.0515 (1) & 0.0624 (2) & \textbf{0.0675} (3) &  & 3 & 89\\
\bottomrule
\end{tabular}

\end{table}

\section{Weight averaging: the validation-selected variant}
\label{app:ema}

An exponential moving average (EMA) of the weights~\cite{polyak1992averaging} during \sast (decay 0.98 per step; the
averaged weights are validated, selected and decoded) reached validation BLEU-4 0.1516 and 0.1539 in the first two seeds
during development, whereas its third seed had been stopped after one epoch (\supp{app:negative}, variant (v)). Before
any test decoding of these checkpoints we fixed, with a time stamp, the following rule (\texttt{ema/DECISION\_RULE.txt}
of the release): the EMA variant replaces \sast if and only if its mean validation BLEU-4 over three seeds exceeds that of
\sast (0.1470); the test split is then decoded once per checkpoint and the result is reported whatever it is. Because the
abstention, per-variable clinical and LUAD analyses had been run with \sast, \method (with \sast) remains the main model
of the paper, and the EMA variant is reported as an additional row of Table~\ref{tab:main} with its own paired tests. The
third seed was then trained to completion with the identical configuration (stopped run kept in the release).
Table~\ref{tab:ema} gives the result: the mean validation BLEU-4 is 0.1507, so the rule holds, and \method-EMA is the
validation-selected variant. On the test split it raises BLEU-4 in every seed, the mean from 0.1425 to 0.1445, and
BLEU-1 and BLEU-2 significantly (Table~\ref{tab:main}). It does not dominate \method: its agreement with the clinical
record is lower (macro-F1 0.279 vs.\ 0.328) and so is its contradiction rate (0.209 vs.\ 0.232), both significant,
because the averaged weights state fewer of the seven variables (53\% vs.\ 63\% of (report, variable) pairs); CRQS is
unchanged (0.267 vs.\ 0.269).

\begin{table}[!htb]
\caption{\textbf{\method-EMA.} Validation BLEU-4 of the selected epoch, and test metrics (test decoded once per
checkpoint, after the rule was applied).}
\label{tab:ema}
\centering\footnotesize
\setlength{\tabcolsep}{4pt}
\begin{tabular}{@{}lcccccccc@{}}
\toprule
 & \multicolumn{2}{c}{Val B-4} & \multicolumn{2}{c}{Test B-4} & \multicolumn{2}{c}{Test MTR} & \multicolumn{2}{c}{Test R-L}\\
\cmidrule(lr){2-3}\cmidrule(lr){4-5}\cmidrule(lr){6-7}\cmidrule(lr){8-9}
Seed & \sast & EMA & \sast & EMA & \sast & EMA & \sast & EMA\\
\midrule
9233 & 0.1463 & 0.1516 & 0.1452 & 0.1471 & 0.1753 & 0.1744 & 0.3042 & 0.3043\\
1 & 0.1480 & 0.1539 & 0.1421 & 0.1455 & 0.1738 & 0.1772 & 0.3009 & 0.3036\\
2 & 0.1468 & 0.1466 & 0.1402 & 0.1408 & 0.1720 & 0.1727 & 0.2969 & 0.2997\\
\midrule
Mean & 0.1470 & \textbf{0.1507} & 0.1425 & \textbf{0.1445} & 0.1737 & \textbf{0.1748} & 0.3007 & \textbf{0.3025}\\
\bottomrule
\end{tabular}
\end{table}

\section{Design variants evaluated on validation}
\label{app:negative}

Every variant below was specified before training, trained with the settings of the model it modifies and selected on
validation BLEU-4 with the same rule as \method; none of them was adopted, and none except VED-copy was evaluated on the
test split. All run folders, logs and checkpoints are part of the release.

\noindent\textbf{Reusing retrieved \emph{content} (VED).} \methodbase compresses every retrieved report into one
max-pooled BERT vector, so its decoder can neither see the phrasing of similar cases nor reuse their statements, and
adaptive knowledge fusion switches this branch off (Sec.~\ref{app:backbone}). Verified Exemplar Distillation replaced the
bottleneck by a token-level memory of the five most image-similar training reports, with every exemplar token verified
against the query slide and the exemplar's own slide, and optionally a copy distribution. Validation BLEU-4 (seed 9233):
\methodbase 0.114, VED 0.101, VED with copying 0.107; the copying variant, evaluated once on test as the better
one, reached 0.108 against \methodbase's 0.117. Only 38.7\% of the finding statements of retrieved reports hold for the query
(Sec.~\ref{sec:analysis}), and the slide evidence separates true from false exemplar statements with AUC 0.65. Retrieved
content is mostly another patient's while retrieved structure transfers, which is why \method takes only the structure
from retrieved cases and the findings from the slide.

\noindent\textbf{\sast without its structure anchor.} In a preliminary study on another feature set,
self-critical training of \methodbase with a sentence-BLEU-4 reward diverged: sampled reports grew to 181 to 187 tokens and
validation BLEU-4 fell from 0.109 to 0.084 and 0.054 in the first two epochs. With the structure signals of \psd, the
samples of \sast stay at the planned length (139 to 141 tokens) for all eight epochs in two seeds and for five in the third,
after which that seed's decoder stopped following the budget (validation selection kept its epoch-4 checkpoint). The
anchor thus moves the length drift of sequence-level training well beyond the best validation checkpoint.

\noindent\textbf{Reward composition.} A pilot of \sast with the reward BLEU-4 + ROUGE-L (seed 9233) raised validation
BLEU-4 from 0.1430 to 0.1483 but lowered METEOR (0.173 to 0.167) and the rule-based CE-F1 (0.438 to 0.409); the reward of
\sast therefore includes the report-level METEOR.

\noindent\textbf{Further variants (Table~\ref{tab:variants}).} (i) A \emph{learned structure planner}, a policy over the
plans and length budgets of the ten most image-similar reports trained jointly with the decoder by REINFORCE, learned to
lengthen the reports (mean validation budget 147 $\rightarrow$ 195 tokens) and lowered validation BLEU-4 in every epoch.
(ii) A \emph{finding-consistency reward}, the $F_{0.5}$ score between the findings of a sampled report and of the
reference added to the reward of \sast, lowered validation BLEU-4 and, although it rewards them directly, the F1 of the
stated findings in every seed and epoch. (iii) \emph{Sentence-level length budgets}, the length of every planned sentence
taken from the exemplar report, lowered BLEU-4 at the cross-entropy stage. (iv) \emph{Aligned exemplar editing}, an
attention and copy distribution over the exemplar sentence aligned to the sentence being written, motivated by the
complementarity of the exemplar's wording (a sentence-wise oracle combination of \method's and the exemplar's sentences
reaches validation BLEU-4 0.164), lowered BLEU-4 at the cross-entropy stage and after sequence-level training,
presumably because copying imports another patient's details together with the format. (v) An exponential moving average of the weights during \sast (decay 0.98 per
step) reached validation BLEU-4 0.1516 and 0.1539 in the first two seeds (\sast: 0.1463 and 0.1480); its third seed was
completed under a pre-registered rule and it is reported as \method-EMA (\supp{app:ema}).

\begin{table}[!htb]
\caption{Design variants: best validation BLEU-4, mean over seeds (number of seeds), against the model each one modifies.}
\label{tab:variants}
\centering\footnotesize
\setlength{\tabcolsep}{5pt}
\begin{tabular}{@{}lccc@{}}
\toprule
Variant & Stage & Variant & Reference model\\
\midrule
(i) learned structure planner & sequence-level & 0.1430 (1) & 0.1430 (\methodce, same seed)\\
(ii) finding-consistency reward & sequence-level & 0.1419 (3) & 0.1470 (\method)\\
(iii) sentence-level length budgets & cross-entropy & 0.1284 (3) & 0.1419 (\methodce)\\
(iv) aligned exemplar editing & cross-entropy & 0.1276 (3) & 0.1419 (\methodce)\\
(iv) aligned exemplar editing & sequence-level & 0.1362 (3) & 0.1470 (\method)\\
\bottomrule
\end{tabular}
\end{table}

\section{Data and settings}
\label{app:config}

Table~\ref{tab:allsettings} lists every setting of the final model; the release contains the resolved configuration,
the source code and the software environment of every run. Settings marked $^{\dagger}$ were chosen on the validation split by a rule
fixed in advance (Sec.~\ref{sec:setup}); all others are standard values of the architecture or of the benchmark, fixed
before training.

\begin{table}[!htb]
\caption{Data. Tokens after the benchmark cleaner; tiles are UNI2-h tiles of 256\,px at 20$\times$.}
\label{tab:data}
\centering\footnotesize
\setlength{\tabcolsep}{2.4pt}
\begin{tabular}{@{}lrrr@{}}
\toprule
 & Train & Val & Test\\
\midrule
Patients (= slides) & 792 & 85 & 92\\
Tokens per report, mean / median & 139.7 / 125 & 149.4 / 133 & 149.2 / 138\\
Sentences per report, mean & 8.9 & 9.4 & 9.5\\
Contributing sites (TSS) & 37 & n/a & n/a\\
Tiles per slide, mean / max & \multicolumn{3}{c}{12{,}016 / 38{,}962}\\
\bottomrule
\end{tabular}
\end{table}

\begin{table}[!htb]
\caption{Settings of \method (and of \methodce and \methodbase, which it contains).}
\label{tab:allsettings}
\centering\footnotesize
\setlength{\tabcolsep}{4pt}
\begin{tabular}{@{}lll@{}}
\toprule
Group & Setting & Value\\
\midrule
Input & tile encoder / tiles & UNI2-h, 1536-d, 256\,px at 20$\times$, all tissue tiles\\
 & bag cap & 10{,}000 tiles (random subset per training epoch; evenly spaced at evaluation)\\
\ske, \pcd & model width / heads / FFN & $d{=}512$ / 8 / 2048\\
 & decoder layers / dropout & 3 / 0.1\\
 & topics $N_T$ / retrieved reports $N_K$ & 20 / 100 (max-pooled BERT embeddings)\\
 & finding graph & 20 topic nodes + global node, 69 edges, 5 tissue compartments, 2 residual graph convolutions\\
 & knowledge fusion & adaptive knowledge fusion, exact per-tile decomposition (Sec.~\ref{app:backbone})\\
Stage 1 & loss / selection & weighted BCE ($w_j{=}n_j^-/n_j^+$), graph-level and per-node classifiers / validation macro-AUC, patience 10\\
 & batch / epochs & 16 patients / at most 100\\
Stage 2 & optimiser & Adam, $\beta{=}(0.8,0.999)$, learning rate $10^{-4}$, no weight decay\\
 & batch / clipping / precision & 16 patients / gradient value 0.1 / bfloat16 autocast\\
 & validation & every 2 epochs from epoch 10; early stop after 20 epochs; monitor BLEU-4\\
 & graph batch-norm & running statistics frozen after stage 1 (scale and shift trained)\\
Decoding & search & beam 3, trigram blocking, no immediate repeats, unknown-word token suppressed, max.\ 600 tokens\\
\psd & sentence types & $C{=}24$ (TF-IDF unigrams and bigrams, min.\ df 3, SVD 64, $\ell_2$, $k$-means, seed 0)\\
 & signals & remaining sentences ($R{=}16$), current type, remaining tokens$^{\dagger}$\\
 & prior structure & medoid plan of $K{=}10$, mean length of $K'{=}5$ most image-similar cases\\
 & plan distance & edit distance divided by the length of the longer plan\\
\ccd & decision & calibrated slide-level posterior $\ge\tau$; $\tau$ = 0.17 / 0.19 / 0.27 (seeds 9233 / 1 / 2)$^{\dagger}$\\
 & claim tokens & topic-bag embedding + $\sqrt{d}\,\times$ present/absent state; one attention per decoder layer\\
\csp & placement & $\arg\max_i P(j\mid\hat\pi_i)$ over the prior plan; training sentences for $P(j\mid k)$\\
 & signal & $\mathbf{V}\hat{\mathbf{a}}_{\sigma_t+1}$ added to the decoder input\\
\sast & initialisation & the validation-selected \methodce checkpoint of the same seed\\
 & reward & sentence-level BLEU-4 + METEOR (pycocoevalcap) + ROUGE-L ($\beta{=}1.2$) against the reference\\
 & estimator & 5 samples per slide (temperature 1, test-time structure and claims), leave-one-out baseline\\
 & optimiser & Adam, learning rate $10^{-5}$, 16 slides per step, gradient value clipping 0.1, no dropout\\
 & selection & validation BLEU-4 every epoch; at most 8 epochs, early stop after 5 without gain\\
Seeds & & 9233, 1, 2 (stage 1 and stage 2 share the seed)\\
\bottomrule
\end{tabular}
\end{table}

\end{document}